%% file: document.tex
\documentclass[journal]{IEEEtran}

\usepackage[utf8]{inputenc}
\usepackage{booktabs}                  
\usepackage[T1]{fontenc}               
\usepackage{glossaries}                
\usepackage{graphicx}
\usepackage{multirow}                  
\usepackage{pgf}
\makeatletter                          
\let\pgfmathModX=\pgfmathMod@          
\usepackage{pgfplots}
\let\pgfmathMod@=\pgfmathModX          
\makeatother
\usepackage{pgfplotstable}             
\usepackage[caption=false]{subfig}
\usepackage{hyperref}
\usepackage{algorithm}
\usepackage{algorithmic}
\usepackage[numbers,square,sort&compress]{natbib}
\usepackage{amsfonts}                  
\usepackage{bm}                        

\usepackage{tikz}
\usepackage{flowchart}
\usetikzlibrary{positioning}      
\usetikzlibrary{shapes.geometric} 
\usetikzlibrary{calc}             
\usetikzlibrary{chains}
\usetikzlibrary{arrows}
\usetikzlibrary{scopes}           
\usetikzlibrary{spy}              
\usetikzlibrary{fit}

\graphicspath{{figure/}}

\pgfplotsset{
	compat=1.12,                   
	tick label style={
		font=\footnotesize
	},
	label style={
		font=\footnotesize
	},
	legend style={
		font=\scriptsize
	},
	every axis/.append style={
		cycle list name=color list,
		thick,
		tick style={semithick},
	},
}

\pgfplotstableset{
	col sep=comma,                 
	every head row/.style={        
		before row=\toprule,   %
		after row=\midrule     %
	},                             %
	every last row/.style={        %
		after row=\bottomrule  %
	},                             %
	fixed,                         
	fixed zerofill,                %
	precision=2,                   %
}

\def\showdetail#1#2#3#4#5#6{%
\node[image,#2] (#1) {\includegraphics[width=\imw]{#1}};
\path (#1.south west) ++(\detail) coordinate (#1 detail position);
\spy on (#1 detail position) in node [#4=of #1.north #5,anchor=north #6];
\node[#4=of #1.south #5, anchor=south #6] {#3};
}

\def\showdetailall#1#2#3#4#5#6{%
\def\imw{#3} 
\def\detail{#2}
\def\hdist{#6}
\begin{tikzpicture}[
	spy using outlines={%
		rectangle,
		magnification = #5,
		size = #4,
		connect spies,
		very thick,
		red,
	},
	node distance = 4mm,
	image/.style = {anchor=south west, inner sep=0},
]
\showdetail{#1_gt}{}{Ground truth}{right}{east}{west}
\showdetail{#1_bicubic}{below=of  #1_gt}{Bicubic}{right}{east}{west}
\showdetail{#1_nbsrf}{below=of    #1_bicubic}{NBSRF}{right}{east}{west}
\showdetail{#1_arflplus}{below=of #1_nbsrf}{ARFL+}{right}{east}{west}
\showdetail{#1_aplus}{below=of    #1_arflplus}{A+}{right}{east}{west}

\showdetail{#1_lr}{right=\hdist of #1_gt}{Low resolution}{left}{west}{east}
\showdetail{#1_srex}{below=of      #1_lr}{SelfEx}{left}{west}{east}
\showdetail{#1_vdsr}{below=of      #1_srex}{VDSR}{left}{west}{east}
\showdetail{#1_drcn}{below=of      #1_vdsr}{DRCN}{left}{west}{east}
\showdetail{#1_icf}{below=of       #1_drcn}{\nameofthegame}{left}{west}{east}
\end{tikzpicture}
}

\def\nameofthegame{\gls{wsdsr}}
\def\nameofthefilter{\gls{wsd}}
\def\mnameofthefilter{\text{WSD}}

\input{acronyms.tex}

\glsunset{aplus}
\glsunset{arflplus}
\glsunset{nbsrf}
\glsunset{vdsr}
\glsunset{drcn}
\glsunset{srex}

\floatname{algorithm}{Procedure}

\begin{document}

\title{Single Image Super-Resolution based on Wiener Filter in Similarity 
Domain}
\author{Cristóvão~Cruz*, 
        Rakesh~Mehta, 
        Vladimir~Katkovnik, 
        and~Karen~Egiazarian \IEEEmembership{Senior~Member,~IEEE}%
\thanks{This work is supported by the Academy of Finland, project no. 287150, 
2015-2019, 
and European Union's H2020 Framework Programme (H2020-MSCA-ITN-2014) 
under grant agreement no. 642685 MacSeNet}%
\thanks{C. Cruz is with Noiseless Imaging Oy, Korkeakoulunkatu 7, 33720 Tampere, Finland and with the Signal Processing Group, Tampere University of Technology, 33720 Tampere, Finland (e-mail: cristovao@noiselessimaging.com)}%
\thanks{R. Mehta is with United Technology Research Centre Ireland, 4th Floor, Penrose Business Center, Penrose Wharf, Cork City, Co. Cork, Republic of Ireland (e-mail: mehtar1@utrc.utc.com)}%
\thanks{V. Katkovnik and K. Egiazarian are with the Signal Processing Group, Tampere University of Technology, 33720 Tampere, Finland (e-mail: vladimir.katkovnik@tut.fi, karen.egiazarian@tut.fi)}%
}


\maketitle

\begin{abstract}

  Single image super resolution (SISR) is an ill-posed problem aiming at
  estimating a plausible high resolution (HR) image from a single low resolution
  (LR) image. Current state-of-the-art SISR methods are patch-based.  They use
  either external data or internal self-similarity to learn a prior for a HR
  image. External data based methods utilize large number of patches from the
  training data, while self-similarity based approaches leverage one or more
  similar patches from the input image. In this paper we propose a
  self-similarity based approach that is able to use large groups of similar
  patches extracted from the input image to solve the SISR problem. We introduce
  a novel prior leading to collaborative filtering of patch groups in 1D
  similarity domain and couple it with an iterative back-projection
  framework. The performance of the proposed algorithm is evaluated on a number
  of SISR benchmark datasets. Without using any external data, the proposed
  approach outperforms the current non-CNN based methods on the tested datasets
  for various scaling factors.  On certain datasets, the gain is over 1 dB, when
  compared to the recent method A+. For high sampling rate (x4) the proposed
  method performs similarly to very recent state-of-the-art deep convolutional
  network based approaches.

\end{abstract}



\section{Introduction}

\IEEEPARstart{T}{he} goal of \gls{sisr} is to estimate the high frequency
spectrum of an image from a single band limited measurement. In other words, it
means generating a plausible high resolution image that does not contradict the
low resolution version used as input. It is a classical problem in image
processing which finds numerous applications in medical imaging, security,
surveillance and astronomical imaging, to name few. Simple methods based on
interpolation (e.g., bilinear, bicubic) are frequently employed because of their
computational simplicity, but due to use of low order polynomials, they mostly
yield very smooth results that do not contain the sharp edges or fine textures,
often present in natural images.

In recent years, these shortcomings have been partially resolved by approaches
that use machine learning to generate a \gls{lr} to \gls{hr} mapping from a
large number of images \cite{timofte_2014_adjusted, schulter_2015_fast}.
Existing methods utilized to learn this mapping include manifold learning
\cite{chang_2004_super}, sparse coding \cite{yang_2010_image}, \glspl{cnn}
\cite{dong_2014_learning, kim_2016_accurate, kim_2016_deeply}, and local linear
regression \cite{timofte_2013_anchored, timofte_2014_adjusted}. The prior
learned by these approaches has been shown to effectively capture natural image
structure, however, the improved performance comes with some strong limitations.
First, they heavily rely on a large amount of training data, which can be very
specific for different kind of images and somehow limits the domain of
application.  Second, a number of these approaches, most markedly the \gls{cnn}
based ones, take a considerable amount of training time, ranging from several
hours to several days on very sophisticated \glspl{gpu}. Third, a separate
\gls{lr}-\gls{hr} mapping must be learned for each individual up-sampling factor
and scale ratio, limiting its use to applications were these are known
beforehand. Finally, a number of these approaches \cite{timofte_2014_adjusted,
timofte_2013_anchored}, do not support non-integer up-sampling factors.

Certain researchers have addressed the \gls{sisr} problem by exploiting the
priors from the input image in various forms of self-similarity
\cite{glasner_2009_super}, \cite{freedman_2011_image}, \cite{cui_2014_deep},
\cite{dong_2013_sparse}.  \citet{freedman_2011_image} observed that, although
fewer in number, the input image based search results in ``more relevant
patches''. Some self-similarity based algorithms find a \gls{lr}-\gls{hr} pair
by searching for the most similar target patch in the down-sampled image
\cite{glasner_2009_super, freedman_2011_image, huang_2015_single,
singh_2014_sub}. Other approaches are able to use several self-similar patches
and couple them with sparsity based approaches, such as
\citet{dong_2013_sparse}.  \citet{yang_2013_self} are also able to self-learn a
model for the reconstruction using sparse representation of image patches.
\citet{shi_2016_lowrank} use a low-rank representation of non-local
self-similar patches extracted from different scales of the input image. These
approaches do not required training or any external data, but their performance
is usually inferior to approaches employing external data, especially on natural
images with complex structures and low degree of self-similarity. Still, in all
of them, sparsity is regarded as an instrumental tool in improving the
reconstruction performance over previous attempts.

In this work we propose \gls{wsdsr}, a technique for \gls{sisr} that
simultaneously considers sparsity and consistency. To achieve this aim, we
formulate the \gls{sisr} problem as a minimization of reconstruction error
subject to a sparse self-similarity prior. The core of this work lies in the
design of the regularizer that enforces sparsity in groups of self-similar
patches extracted from the input image. This regularizer, which we term
\gls{wsd}, is based on \gls{bm3d} \citep{dabov_2007_color,dabov_2007_image}, but
includes particular twists that make a considerable difference in \gls{sisr}
tasks. The most significant one is the use of a 1D Wiener filter that only
operates along the dimension of similar patches. This feature alone, mitigates
the blur introduced by the regularizers designed for denoising that make use of 3D
filtering and proved essential for the high performance of our proposed method (see
Table~\ref{tab:transform}).


\section{Contribution and Structure of the Paper}
\label{sec:contribution}

The main characteristics of the proposed approach are as follows: 

\begin{enumerate}
\item \emph{No external data or training required:} the proposed approach
exploits the image self-similarity, therefore, it does not require any external
data to learn an image prior, nor does it need any training stage;
\item \emph{Supports non-integer scaling factors:} the image can be scaled by any
factor and aspect ratio;
\item \emph{No border pruning effect:} The proposed approach represents the
complete image in the high resolution space without any border pruning effect,
unlike most of the dictionary based algorithms \cite{zeyde_2012_single,
timofte_2014_adjusted}.
\item \emph{Excellent performance:} it competes with the state-of-the-art
approaches in both computational complexity and estimation quality as will be
demonstrated in section~\ref{sec:experiments}.
\end{enumerate}

The previous conference publication of the proposed approach was done in
\cite{egiazarian_2015_single}. The algorithm in this paper follows the general
structure of \cite{egiazarian_2015_single}, but introduces a novel regularizer
that proved crucial for obtaining significantly improved performance. The
distinctive features of the developed algorithm are:

\begin{itemize}
\item 1D Wiener filtering along similarity domain;
\item Reuse of grouping information;
\item Adaptive search window size;
\item Iterative procedure guided by input dependent heuristics;
\item Improved parameter tuning.
\end{itemize}

An extensive simulation study demonstrates the advanced performance of the
developed algorithm as compared with \cite{egiazarian_2015_single} and some
state-of-the-art methods in the field.

The paper is organized as follows. In Section~\ref{sec:related} we provide an
overview of modern single image super-resolution methods. In
Section~\ref{sec:framework} we formulate the problem and present the framework
we used to solve it. Section~\ref{sec:filter} contains the main contribution of
this paper and provides a detailed exposition and analysis of the novel
regularizer to be employed within the presented
framework. Section~\ref{sec:experiments} provides an experimental analysis of
our proposal and comparison against several other \gls{sisr} methods, both
quantitative and qualitative.  Section~\ref{sec:discussion} analyses possible
variations of the proposed approach that could lead to further improvements.
Finally, Section~\ref{sec:conclusion} provides a summary of the work.

\section{Related Work}
\label{sec:related}

The \gls{sisr} algorithms can be broadly divided into two main classes: 
the methods that rely solely on observed data and those  that additionally use external data. 
Both of these classes can be further divided into the following categories: learning-based and reconstruction-based. However, we are going to present below the related work in a simplified division of the methods that only accounts for use, or lack of use, of external data without any aim to be considered as an extensive review of the field.

\subsection{Approaches Using External Data}

This type of approaches use a set of \gls{hr} images and their down-sampled
\gls{lr} versions to learn dictionaries, regression functions or end-to-end
mapping between the two. Initial dictionary-based techniques created a
correspondence map between features of \gls{lr} patches and a single \gls{hr}
patch \citep{freeman_2002_example}. Searching in this type of dictionaries was
performed using \gls{ann}, as exhaustive search would be prohibitively
expensive. Still, dictionaries quickly grew in size with the amount of used
training data. \citet{chang_2004_super} proposed the use of \gls{lle} to better
generalize over the training data and therefore require smaller
dictionaries. Image patches were assumed to live in a low dimensional manifold
which allowed the estimation of high resolution patches as a linear combination
of multiple \emph{nearby} patches.  \citet{yang_2010_image} also tackled to
problem of growing dictionary sizes, but using sparse coding. In this case, a
technique to obtain a sparse ``compact dictionary'' from the training data is
proposed. This dictionary is then used to find a sparse activation vector for a
given \gls{lr} patch. The \gls{hr} estimate is finally obtained by multiplying
the activation vector by the \gls{hr} dictionary.
\citet{yang_2012_coupled,zeyde_2012_single} build on this approach and propose
methods to learn more compact dictionaries. \citet{ahmed_2016_single} learns
multiple dictionaries, each containing features along a different direction. The
high-resolution patch is reconstructed using the dictionary that yields the
lowest sparse reconstruction error. \citet{kim_2016_discrete} does away with the
expensive search procedure by using a new feature transform that is able to
perform simultaneous feature extraction and nearest neighbour
identification. Dictionaries can also be leveraged together with regression based
techniques to compute projection matrices that, when applied to the \gls{lr}
patches, produce a \gls{hr} result. The papers by
\citet{timofte_2013_anchored,timofte_2014_adjusted,timofte_2016_improved} are
examples of such an approach where for each dictionary atom, a projection matrix
that uses only the nearest atoms is computed. Reconstruction is performed by
finding the nearest neighbour of the \gls{lr} patch and employing the
corresponding projection matrix. \citet{zhang_2016_joint} follows a similar
approach but also learns the clustering function, reducing the required amount
of anchor points. Other approaches do not build dictionaries out of the training
data, but chose to learn simple operators, with the advantage of creating more
computationally efficient solutions. \citet{tang_2017_pairwise} learns two small
matrices that are used on image patches as left and right multiplication
operator and allow fast recovery of the high resolution image. The global nature
of these matrices, however, fails to capture small details and complex textures.
\citet{choi_2017_single} learns instead multiple local linear mappings and a
global regressor, which are applied in sequence to enforce both local and global
consistency, resulting in better representation of local structure.
\citet{sun_2011_gradient} learns a prior and applies it using a conventional
image restoration approach. Finally, neural networks have also been explored to
solve this problem, in various ways. \citet{sidike_2017_fast} uses a neural
network to learn a regressor that tries to follow
edges. \citet{zeng_2017_coupled} proposes the use of \gls{cda} to learn both
efficient representations for low and high resolution patches as well as a
mapping function between them. However, a more common use of this type of
computational model is to leverage massive amounts of training data and learn a
direct low to high resolution image mapping
\citep{dong_2016_image,liu_2016_robust,kim_2016_accurate,kim_2016_deeply}. Of
these approaches, only \citet{liu_2016_robust} tries to include domain expertise
in the design phase, and despite the fact that testing is relatively
inexpensive, training can take days even on powerful computers.

Although these approaches learn a strong prior from the large amount of training
data, they require a long time to train the models. Furthermore, a separate
dictionary is trained for each up-sampling factor, which limits the available
up-sampling factors during the test time.

\subsection{Approaches Based Only on Observed Data}

This type of approaches rely on image priors to generate an \gls{hr}
image having only access to the \gls{lr} observation.  Early techniques of
this sort are still heavily used due to their computational simplicity, but the
low order signal models that they employ fail to generate the missing high
frequency components, resulting in over-smoothed estimates.
\citet{haris_2016_first} manages to partially solve this problem by using linear
interpolators that operate only along the edge direction.
\citet{wei_2016_freshfri} explores the use \gls{fri} to enhance linear
up-scaling techniques with piece-wise polynomial estimates. Other solutions use
separate models for the low-frequency and high-frequency components, the smooth areas and
the textures and edges \citep{deng_2016_single,yao_2017_blending}. An
alternative approach to image modeling draws from the concept of
self-similarity, the idea that natural images exhibit high degree of repetitive
behavior. \citet{ebrahimi_2007_solving} proposed a super-resolution algorithm
by exploiting the self-similarity and the fractal characteristic of the image at
different scales, where the non-local means \cite{buades_2005_non} is used to
perform the weighting of patches. \citet{freedman_2011_image} extended the idea
by imposing a limit on the search space and, thereby, reduced the complexity.
They also incorporated incremental up-sampling to obtain the desired image
size. \citet{suetake_2008_image} utilized the self-similarity to generate an
example code-book to estimate the missing high-frequency band and combined it
with a framework similar to \cite{freeman_2002_example}.
\citet{glasner_2009_super} used self-examples within and across multiple image
scales to regularize the otherwise ill-posed classical super-resolution scheme.
\citet{singh_2014_super} proposed an approach for super-resolving the image in
the noisy scenarios. \citet{egiazarian_2015_single}, introduced the sparse
coding in the transform domain to collectively restore the local structure in
the high resolution image. \citet{dong_2013_sparse} also employs self-similarity
to model each pixel as a linear combination of its non-local
neighbors. \citet{cui_2014_deep} utilized the self-similarity with a cascaded
network to incrementally increase the image resolution. Recently,
\citet{huang_2015_single} improved the search strategy by considering affine
transformations, instead of translations, for the best patch match.  Further,
various search strategies have been proposed to improve the \gls{lr}-\gls{hr}
pair based on textural pattern \cite{singh_2014_sub}, optical flow
\cite{zhu_2014_single} and geometry \cite{fernandez-granda_2013_super}.



\section{Framework for Iterative SISR}
\label{sec:framework}

A linear ill-posed inverse problem, typical for image restoration, in
particular, for image deblurring and super-resolution, is considered here for
the noiseless case:
\begin{equation}
\bm{y}=\bm{Hx}\label{1}
\end{equation}
where $\bm{y}\in\mathbb{R}^{m}$, $\bm{x}\in\mathbb{R}^{n}$, $m\leq n$, 
$\bm{H}$ is a known linear operator. 

The problem is to solve (\ref{1}) with respect to $\bm{x}$ provided some prior
information on $\bm{x}$. In terms of super-resolution, the operations in
$\mathbb{R}^{m}$ and $\mathbb{R}^{n}$ can be treated as operations with low- and
high-resolution images, respectively. Iterative algorithms to estimate $\bm{x}$
from (\ref{1}) usually include both up-sampling and down-sampling operations along
with some prior information on these variables.

In this work we solve this inverse problem using a general approach similar to one introduced in \citet{danielyan_2012_bm3d} for image deblurring, but focusing on the specific problem of \gls{sisr}.

The sparse reconstruction of $\bm{x}$ can be formulated as the following
constrained optimization:

\begin{eqnarray}
&&\min_{\bm{\theta}\in\mathbb{R}^{m}}\text{ }||\bm{\theta }||_{0}\text{, }%
||\bm{y}-\bm{Hx||}_{2}^{2}\leq\varepsilon^{2}\text{,}\label{SR_1}\\
&&\text{ }\bm{x}=\bm{\Psi\theta}\text{, }\bm{\theta =\Phi x}.
\nonumber
\end{eqnarray}

Here $\bm{\Phi}\in\mathbb{R}^{m\times n}$ and
$\bm{\Psi}\in\mathbb{R}^{n\times m}$ are analysis and syntheses matrices,
$\bm{\theta}\in\mathbb{R}^{m}$ is a spectrum vector, and $ \varepsilon $ is
a parameter controlling the accuracy of the equation (\ref{1}). For the
super-resolution problem $m<n$. Recall that $l_{0}$-pseudo norm,
$||\bm{\theta }||_{0}$, is calculated as a number of non-zero elements of
$\bm{\theta }$ and $||.||_2$ denotes the $l_2$ norm.

Sparse reconstruction of $\bm{x}$ means minimization of $||\bm{\theta}||_{0}$
corresponding to a sparse representation for $\bm{x}$ provided equations linking
the image with the spectrum and the inequality defining the accuracy of the
observation fitting. While the straightforward minimization (\ref{SR_1}) is
possible, our approach is essentially different. Following
\citet{danielyan_2012_bm3d}, we apply the multiple-criteria Nash equilibrium
technique using the following two cost functions:
\begin{IEEEeqnarray}{rCl}
&&J_{1}(\bm{\theta ,x})=\frac{1}{2\varepsilon ^{2}}||\bm{y}-\bm{Hx}
||_{2}^{2}+\frac{1}{2\gamma }||\bm{x}-\bm{\Psi \theta }||_{2}^{2},
\label{MultCrit1} \\
&&J_{2}(\bm{\theta ,x})=\frac{1}{2}||\bm{\theta }-\bm{\Phi x}
||_{2}^{2}+\tau _{\theta }||\bm{\theta }||_{0}.  \label{MultCrit2}
\end{IEEEeqnarray}
The first summand in $J_{1}$ corresponds to the given observations \ and the
second one is penalization of the equation $\bm{x}=\bm{\Psi \theta }$.
The criterion $J_{2}$ enables the sparsity of the spectrum $\bm{\theta}$ 
for $\bm{x}$ provided the restriction $\bm{\theta = \Phi x}$. The
Nash equilibrium for (\ref{MultCrit1})-(\ref{MultCrit2}) is a consensus of
restrictions imposed by $J_{1}$, $J_{2}$. It is defined as a fixed point 
($\bm{\theta }^{\ast }$, $\bm{x}^{\ast }$) such that:
\begin{IEEEeqnarray}{rCl}
\bm{x}^{\ast} & = & \arg\min_{\bm{x}} J_{1}(\bm{\theta}^{\ast}, 
\bm{x}) \text{,}\\
\bm{\theta}^{\ast} & = & \arg\min_{\bm{\theta}}J_{2} 
(\bm{\theta,x}^{\ast})\text{.}
\end{IEEEeqnarray}
The equilibrium ($\bm{\theta }^{\ast }$, $\bm{x}^{\ast }$) means
that any deviation from this fixed point results in increasing of at least
one of the criteria.

The iterative algorithm looking for the fixed point has the following
typical iterative form \citep{facchinei_2007_generalized}:
\begin{IEEEeqnarray}{rCl}
\bm{x}^{k+1} & = & \arg\min_{\bm{x}}J_{1} 
(\bm{\theta}^{k}, \bm{x}) \text{,}\\
\bm{\theta}^{k+1} & = & \arg\min_{\bm{\theta}}J_{2} 
(\bm{\theta,x}^{k+1})\text{.}
\end{IEEEeqnarray}

We modify this procedure by replacing the minimization of $J_{1}$ on $\bm{x}$
by a gradient descent step corresponding to the gradient 
\begin{equation}
\partial J_{1} / \partial \bm{x}  = \frac{1}{\varepsilon^{2}} \bm{H}^{T}
(\bm{y} - \bm{Hx}) - \frac{1}{\gamma}
(\bm{x} - \bm{\Psi \theta})\text{.}
\label{gradient}
\end{equation}

Accompanied by minimization of $J_{2}$ on $\bm{\theta }$ it gives the
following iterations for the solution of the problem at hand:%
\begin{IEEEeqnarray}{rCl}
\bm{x}^{k+1} & = & \bm{\tilde{x}}^{k}\\
&& +\:\alpha(\bm{H}^{T}\bm{H})^{+}
\left[\frac{1}{\varepsilon^{2}} \bm{H}^{T} (\bm{y} - \bm{H\tilde{x}}^{k}) -
\frac{1}{\gamma} (\bm{x}^{k} - \bm{\tilde{x}}^{k})\right],\nonumber\\
\bm{\theta }^{k+1} & = &\arg\min_{\bm{\theta}}J_2(\bm{\theta ,x}
^{k+1}),\label{ITER1}\\
\bm{\tilde{x}}^{k+1} & = & \bm{\Psi \theta}^{k+1}.
\end{IEEEeqnarray}

Here ($^{+}$) stands for the Moore-Penrose pseudo-inverse.
The matrix $(\bm{H}^{T}\bm{H})^{+}$ is a typical factor used for
acceleration of the gradient iterations.

The optimization on $\bm{\theta }$ in (\ref{ITER1}) gives as a solution
the \gls{ht} with the threshold equal to $\sqrt{2\tau _{\theta }}$, 
where $\tau _{\theta }$ denotes the threshold parameter such that:
\begin{equation}
\bm{\theta }^{k}=Th_{\sqrt{2\tau _{\theta }}}(\bm{\Phi x}^{k})\text{.%
}  \label{ITER11}
\end{equation}

This thresholded spectrum combined the with the equality $\bm{\tilde{x}}^{k} =
\bm{\Psi \theta}^{k}$ defines the filter with input $\bm{x}^{k}$, output
$\bm{\tilde{x}}^{k}$ and threshold parameter $\tau_{\theta}$:
\begin{equation}
\bm{\tilde{x}}^{k}=\mathcal{F}_{1}(\bm{x}^{k},\tau _{\theta })%
\text{.}  \label{F1}
\end{equation}

Then the algorithm can be written in the following compact form 
\begin{IEEEeqnarray}{rCl}
\bm{x}^{k+1}&=&\bm{\tilde{x}}^{k}\label{ITER111}\\
&&+\:\alpha(\bm{H}^{T}\bm{H})^{+}\left[
\bm{H}^{T}(\bm{y}-\bm{H\tilde{x}}^{k})\frac{1}{\varepsilon ^{2}}-\frac{1}{
\gamma}(\bm{\tilde{x}}^{k}-\bm{x}^{k})\right],\nonumber\\
\bm{\tilde{x}}^{k+1}&=&\mathcal{F}_{1}(
\bm{x}^{k+1},\tau_{\theta})\label{eqn:compact_filter}.
\end{IEEEeqnarray}

The first line of this algorithm defines an update of the super-resolution image
$\bm{x}^{k+1}$ obtained from the low resolution residue $\bm{y}
- \bm{H\tilde{x}}^{k}$. Note, that
$(\bm{H}^{T}\bm{H})^{+}\bm{H}^{T}\in\mathbb{R}^{n\times m}$ is an up-sampling
operator (matrix), which we will denote by 
$\bm{U}\in\mathbb{R}^{n\times m}$:
\begin{equation}
\bm{U} = (\bm{H}^{T}\bm{H})^{+}\bm{H}^{T}.
\end{equation}

The last summand $(\bm{\tilde{x}}^{k}-\bm{x}^{k})/\gamma $ in (\ref{ITER111}) is
the scaled difference between the image estimate after and before filtering.
Experiments show that this summand is negligible. Dropping this term, replacing
$\alpha /\varepsilon ^{2}$ by $\alpha$ and exchanging the order of the
operations (\ref{ITER111}) and (\ref{eqn:compact_filter}) we arrive to the
simplified version of the algorithm:
\begin{IEEEeqnarray}{rCl}
\bm{\tilde{x}}^{k} & = &\mathcal{F}_{1}(\bm{x}^{k-1}, \tau_{\theta}),\\
\bm{x}^{k} &=&\bm{\tilde{x}}^{k}+\alpha \bm{U}(\bm{y} - \bm{H\tilde{x}}^{k}).
\label{ITERIV}
\end{IEEEeqnarray}

The filter $\mathcal{F}_{1}$ is completely defined by the used analysis
and synthesis operators $\bm{\Phi }\in\mathbb{R}^{m\times n}$ and 
$\bm{\Psi }\in\mathbb{R}^{n\times m}$. In particular, if the BM3D
block-matching is used for design of the analysis and synthesis operators, the
filter $\mathcal{F}_{1}$ is the BM3D \gls{ht} algorithm (see
\citet{danielyan_2012_bm3d}).

We replaced this BM3D \gls{ht} algorithm 
by our
proposed regularizer \nameofthefilter, which uses the \gls{bm3d} grouping but is
especially tailored for super-resolution problems. 

The algorithm takes now the final form, which is used in our
demonstrative experiments:
\begin{IEEEeqnarray}{rCl}
\bm{\tilde{x}}^{k} & = &\mnameofthefilter(\bm{x}^{k-1}, \tau^{k}_{\theta}),\label{ITER1V}\\
\bm{x}^{k} &=&\bm{\tilde{x}}^{k}+\alpha \bm{U}(\bm{y} - \bm{H\tilde{x}}^{k}) \text{.}
\end{IEEEeqnarray}

This proposed iterative algorithm is termed \textbf{\gls{wsdsr}}, and formally
described in Procedure~\ref{pro:wsdsr}.

\begin{algorithm}
\caption{WSD-SR algorithm\label{pro:wsdsr}}
\begin{algorithmic}[1]
	\REQUIRE{$y$: low resolution input}
	\REQUIRE{$H$: sampling operator}
	\REQUIRE{$K$: number of iterations}
	\ENSURE{High resolution estimate}

	\STATE $U = (H^TH)^+H^T$ \COMMENT{up-sampling matrix}
	\STATE $x^0 = Uy$
	\COMMENT{initial estimate}
	\FOR{$k=1$ \TO K}
		\STATE $\tilde{x}^k$ = WSD($x^{k-1}$, $\tau_{\theta}^k$)
		\STATE $x^k = \alpha U(y - H\tilde{x}^k) + \tilde{x}^k$
	\ENDFOR

	\RETURN $x^K$
\end{algorithmic}
\end{algorithm}

Note, that (\ref{ITER1V}) defines the regularizing stage of the super-resolution
algorithm. The analysis and synthesis transforms used in \nameofthefilter\ are
data dependent and, as a result, vary from iteration-to-iteration, as in BM3D. The
parameter $\tau_{\theta}$ is also changing throughout the iterations, in order
to account for the need to reduce the strength of the regularizer in the later
stages of the iterative procedure.
 

\section{Proposed Regularizer}
\label{sec:filter}

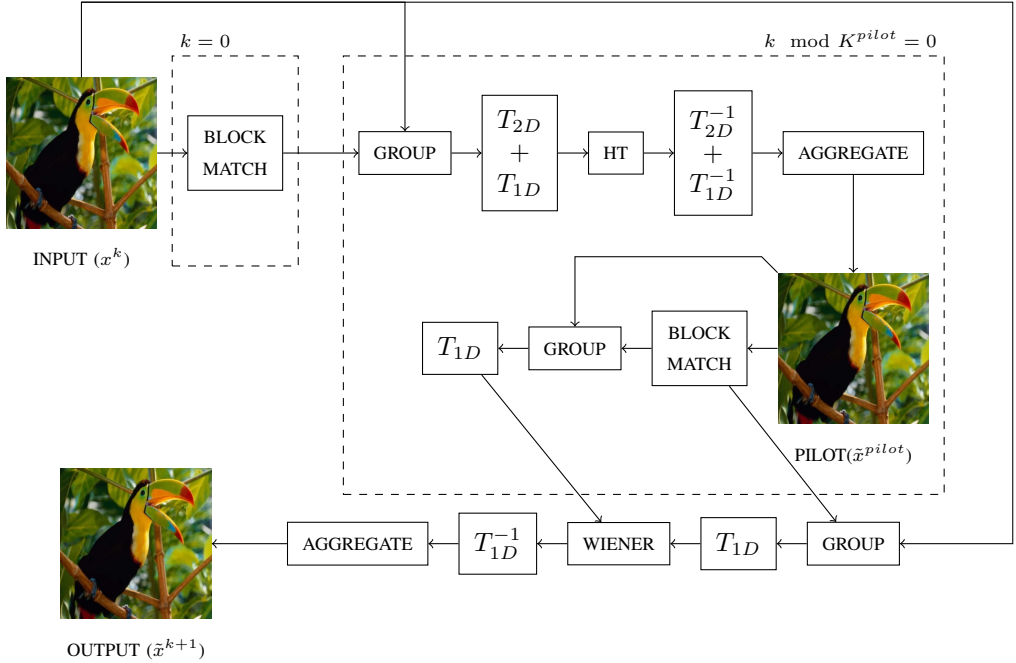
\begin{figure*}
\centering
\input{figure/flowchart.tikz}
\caption{\nameofthefilter\ block diagram.}
\label{fig:flowchart}
\end{figure*}

The proposed regularizer, \gls{wsd}, is highly influenced by the \gls{bm3d}
collaborative filtering scheme that explores self-similarity of natural images
\citep{dabov_2007_image}.

\begin{algorithm}
\caption{WSD algorithm\label{pro:wsd}}
\begin{algorithmic}[1]
	\REQUIRE{$x$: filter input}
	\REQUIRE{$\tau_{\theta}$: filter strength}
	\REQUIRE{$K^{pilot}$: pilot recompute period}
	\REQUIRE{$k$: current iteration}
	\ENSURE{$\tilde{x}$: estimate}

	\COMMENT{Compute match table for pilot estimation.}
	\IF{$k=0$}
		\STATE $m^{ht} \leftarrow $ HTBlockMatch $(x)$
	\ELSE
		\STATE $m^{ht} \leftarrow m^{ht}_{previous}$
	\ENDIF

	\STATE{}
	\STATE{}
	\COMMENT{Pilot estimation.}
	\IF{$k \mod K^{pilot} = 0$}
		\STATE $g^{ht} \leftarrow $ Group $(x, m^{ht})$
		\STATE $\tilde{g}^{ht} \leftarrow $ HardThresholding $(g^{ht}, \tau_{\theta})$
		\STATE $\tilde{x}^{pilot} \leftarrow $ Aggregate $(\tilde{g}^{ht})$
		\STATE $m^{pilot} \leftarrow $ WienerBlockMatch $(\tilde{x}^{pilot})$
	\ELSE
		\STATE $\tilde{x}^{pilot} \leftarrow \tilde{x}^{pilot}_{previous}$
		\STATE $m^{pilot} \leftarrow m^{pilot}_{previous}$
	\ENDIF

	\STATE{}
	\STATE{}
	\COMMENT{Filter the input image using pilot information.}
	\STATE $g^{pilot} \leftarrow $ Group $(\tilde{x}^{pilot}, m^{pilot})$
	\STATE $W \leftarrow $ EstimateWiener $(g^{pilot}, \tau_{\theta})$
	\STATE $g^{wiener} \leftarrow $ Group $(x, m^{pilot})$
	\STATE $\tilde{g}^{wiener} \leftarrow $ WienerFilter $(g^{wiener}, W)$
	\STATE $\tilde{x} \leftarrow $ Aggregate $(\tilde{g}^{wiener})$

	\STATE{}
	\STATE{}
	\COMMENT{Store information for future reuse.}
	\STATE $m^{ht}_{previous} \leftarrow m^{ht} $
	\STATE $\tilde{x}^{pilot}_{previous} \leftarrow \tilde{x}^{pilot} $
	\STATE $m^{pilot}_{previous} \leftarrow m^{pilot}$

	\STATE{}
	\RETURN $\tilde{x}$
\end{algorithmic}
\end{algorithm}

As shown in Fig.~\ref{fig:flowchart} and further described in
Procedure~\ref{pro:wsd}, \gls{wsd} operates in two sequential stages, both
filtering groups of similar patches, as measured using the Euclidean
distance. The result of each stage is created by placing the filtered patches
back in their original locations and performing simple average for pixels with
more than one estimate. The two stages employ different filters on the patch
groups. The first stage, which is producing a pilot estimate used by the second
stage, uses \gls{ht} in the 3D transform domain. The second stage on the other
hand, which is generating the final result, uses the result of the first stage
to estimate an empirical Wiener filter in the 1D transform domain, operating
only along the inter-patch dimension, which we call the similarity domain. This
filter is then applied to the original input data.


The use of the 1D Wiener filter in the second stage sets this approach apart
from both \citet{egiazarian_2015_single} and \citet{wang_2016_return}. It
allowed to not only achieve much sharper results and clearer details, but also
reduce the computational cost. Furthermore, the employed grouping procedure
includes two particular design elements that further improved the system's
performance and reduced its computational complexity: reuse of block match
results and adaptive search window size. Finally, as described in the previous
section, \gls{wsd} is applied iteratively in what we term \gls{wsdsr}. This
requires the modulation of the filtering strength in such a way that it is
successively decreased as the steady-state is approached, in a sort of simulated
annealing fashion \citep{guleryuz_2006_nonlinear_a}. We present input dependent
heuristics for the selection of both the minimum number of iterations and the
filter strength curve.

Overall, the main features of our proposal are:

\begin{enumerate}
	\item Wiener filter in similarity domain;
	\item stateful operation with grouping information reuse;
	\item adaptive search window size;
	\item input dependent iterative procedure parameters.
\end{enumerate}

These design decisions, as well as the parameters selection are studied in
this section. Empirical evidence is presented for each decision, both in terms
of reconstruction quality (PSNR) and computational complexity (speed-up factor).
The tests were conducted on \emph{Set5} \cite{bevilacqua_2012_low} using a scale
factor of $4$, and sampling operator $H$ set to bicubic interpolation with
anti-aliasing filter. In all tables, only the feature under analysis changes
between the different columns and the column marked with a * reflects the final
design.

\subsection{Wiener Filter in Similarity Domain}

The original work on collaborative filtering \cite{dabov_2007_image} addresses
the problem of image denoising, hence, exploits not only the correlation between
similar patches but also between pixels of the same patch. It does so by
performing 3D Wiener filtering on groups of similar patches. The spectrum of
each group is computed by a separable 3D transform composed of a 2D spatial
transform $T_{2D}$ and a 1D transform $T_{1D}$ along the similarity
dimension. However, when dealing with the problem of noiseless super-resolution,
employing a 3D Wiener filter results in spatial smoothing, which is further
exacerbated by the iterative nature of the algorithm. In order to avoid this
problem we use $T_{2D} = I$, which means performing 1D Wiener filtering along
the inter-patch similarity dimension. More specifically, given a match table
$m$, a pilot estimate $\tilde{x}^{pilot}$, and an operation $x(:,m)$ that
extracts from $x$ the patches addressed by $m$ as columns, a 1D empirical Wiener
filter $W$ of strength $\tau_{theta}$ is estimated as follows:
\begin{IEEEeqnarray}{rCl}
  \bm{g}^{pilot} & = & \tilde{\bm{x}}^{pilot}(:,m)\\
  \bm{G}^{pilot} & = & \bm{g}^{pilot} \bm{T}_{1D}\\
  \bm{W} & = & \frac{|\bm{G}^{pilot}|^2}{(|\bm{G}^{pilot}|^2 + \tau_{theta}^2)}
\end{IEEEeqnarray}

The filter is applied by performing point-wise multiplication with the spectrum
of the group of similar patches extracted from the input image $x$, using the
same match table $m$ that was used to estimate the Wiener coefficients $W$:
\begin{IEEEeqnarray}{rCl}
  \bm{g}^{wiener} & = & \bm{x}(:,m)\\
  \bm{G}^{wiener} & = & \bm{g}^{wiener} \bm{T}_{1D}\\
  \tilde{\bm{G}}^{wiener} & = & \bm{W} .* \bm{G}^{wiener}\\
  \tilde{\bm{g}}^{wiener} & = & \bm{T}^{-1}_{1D}\tilde{\bm{G}}^{wiener}
\end{IEEEeqnarray}

The resulting filtered group of patches $\tilde{\bm{g}}^{wiener}$ is ready to be aggregated.

These operations are presented in Procedure~\ref{pro:wsd} using symbolic
names. There, the Group() operation stands for $x(:,m)$, EstimateWiener()
stands for equations (23)-(24) and WienerFilter() stands for equations (26)-(28).

Besides dramatically improving the reconstruction quality, this feature
significantly reduces the computational complexity of \gls{wsd} when compared to
a 3D transform based approach, as suggested by the empirical evidence in
Table~\ref{tab:transform}.

\begin{table}[t]
\centering
\caption{Wiener filter in similarity domain effect on performance (\emph{Set5},
x4), using Haar transform as $T_{1D}$. Speedup is a factor relative to $T_{2D}^{wiener}$ = DCT.}
\label{tab:transform}
\input{table/transform.tex}

\end{table}

\subsection{Grouping Information Reuse}

In the proposed approach, we apply collaborative filtering iteratively on the
input image. However, because the structure of the image does not change
significantly between iterations, the set of similar patches remains fairly
constant. Therefore, we decided to perform block matching sparsely and reuse the
match tables. We observed that in doing so, we not only gain in terms of reduced
computational complexity, but also in terms of reconstruction quality. We
speculate that the improved performance stems from the fact that by using a set
of similar patches for several iterations we avoid oscillations between local
minima, and by revising it sporadically, we allow for small structural changes
that reflect the contribution of the estimated high frequencies.

Each iteration of the collaborative filter typically requires the execution of
the grouping procedure twice, the first time to generate the grouping for
\gls{ht} and the second one to generate the grouping for Wiener filtering. We
observed that this iterative procedure is fairly robust to small changes on the
grouping used for the \gls{ht} stage, to the point that optimal results are
achieved when that match table is computed only once. The same is not true for
the Wiener stage's match table, which still needs to be computed every few
iterations, $K^{pilot}$ in Procedure~\ref{pro:wsd}.  Table~\ref{tab:skip}
presents the empirical evidence concerning these observations.

\begin{table}[t]
\centering
\caption{Match table reuse effect on performance (\emph{Set5}, x4). $K^{pilot} =
5$. Speedup is a factor relative to match table reuse: disabled.}
\label{tab:skip}
\input{table/skip.tex}
\end{table}

\subsection{Adaptive Search Window Size}
\label{sec:search}

\begin{figure*}
\centering
\subfloat[Global]{\includegraphics[width=.3\linewidth]{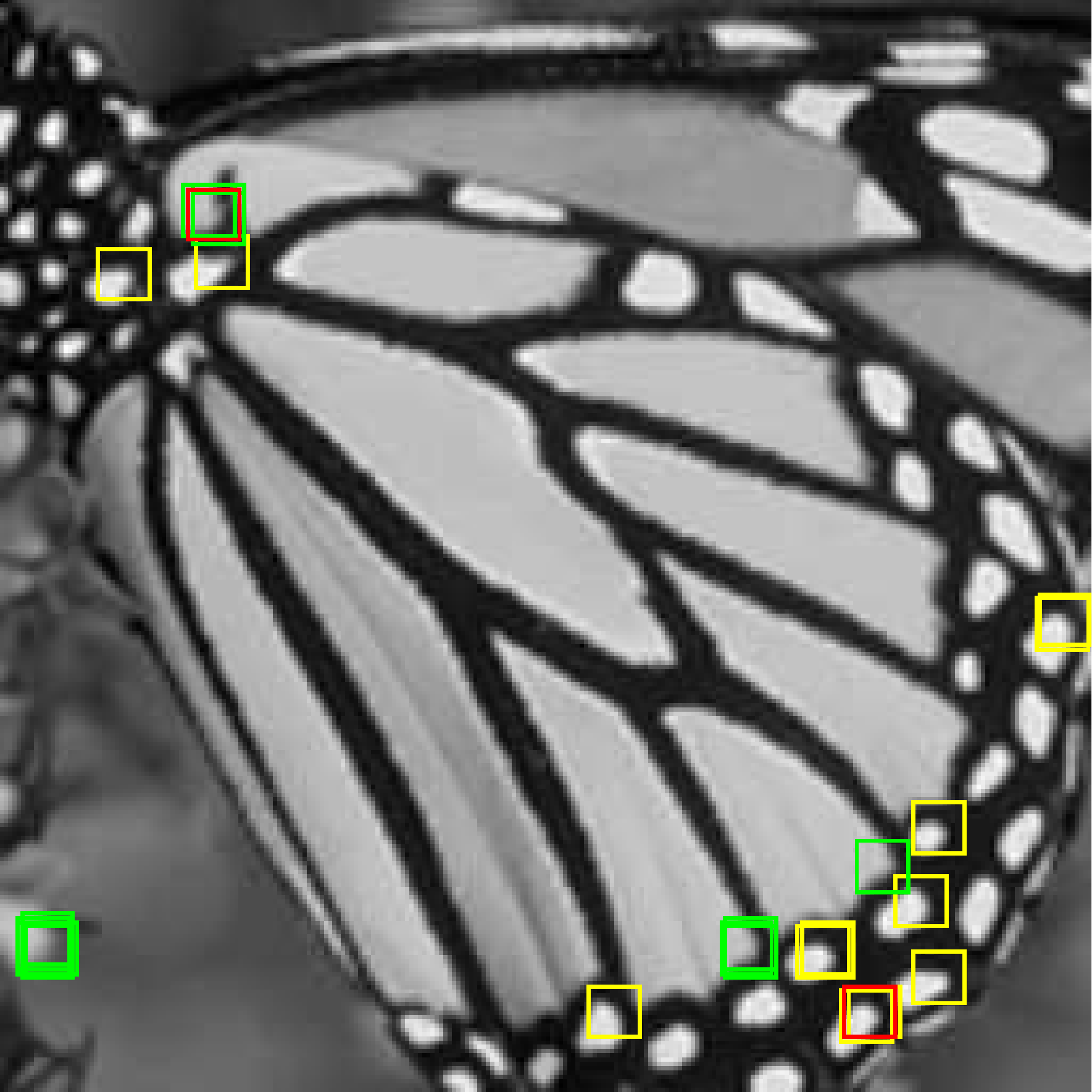}%
\label{fig:search:global}}
\hspace{5mm}
\subfloat[Local]{\includegraphics[width=.3\linewidth]{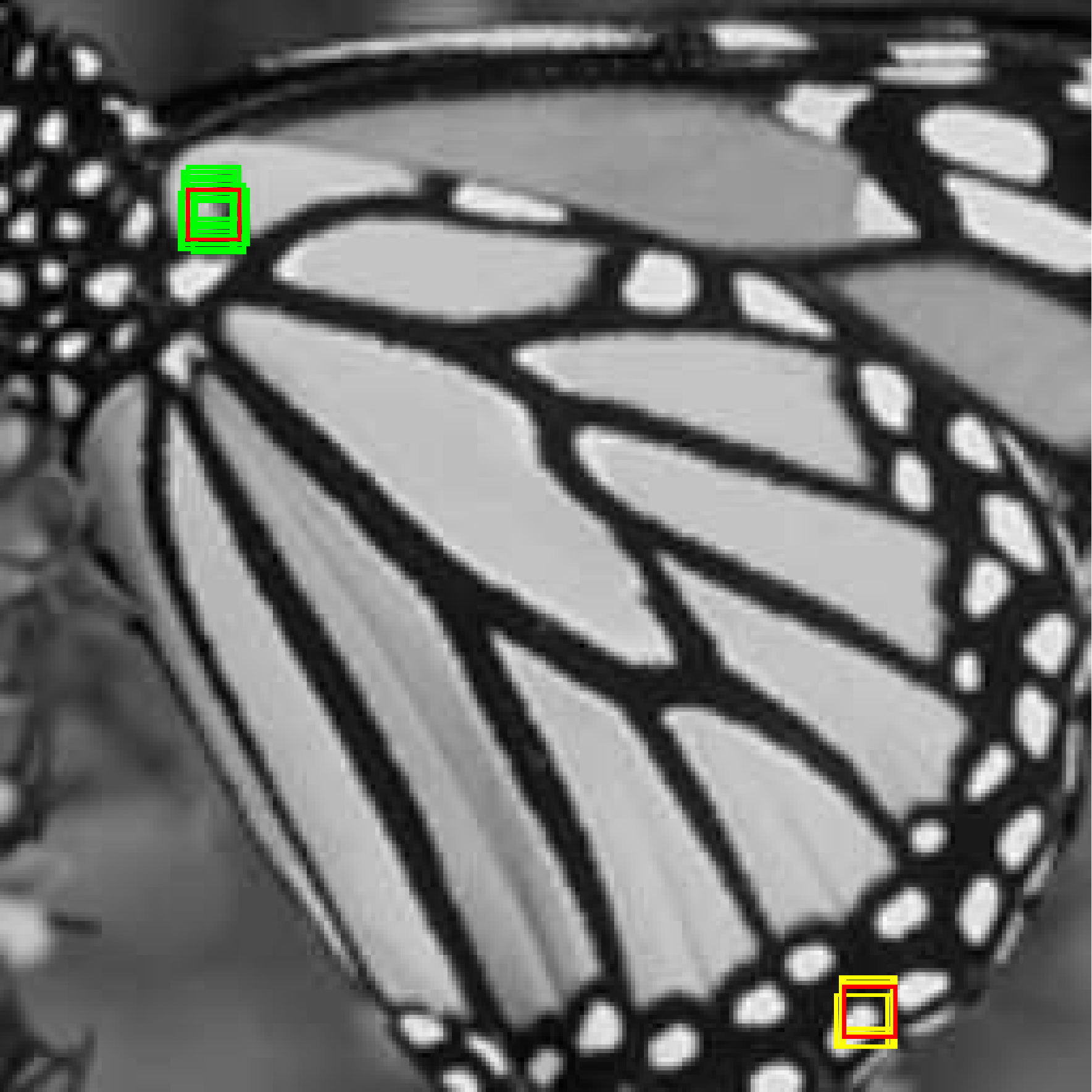}%
\label{fig:search:local}}
\hspace{5mm}
\subfloat[Incremental]{\includegraphics[width=.3\linewidth]{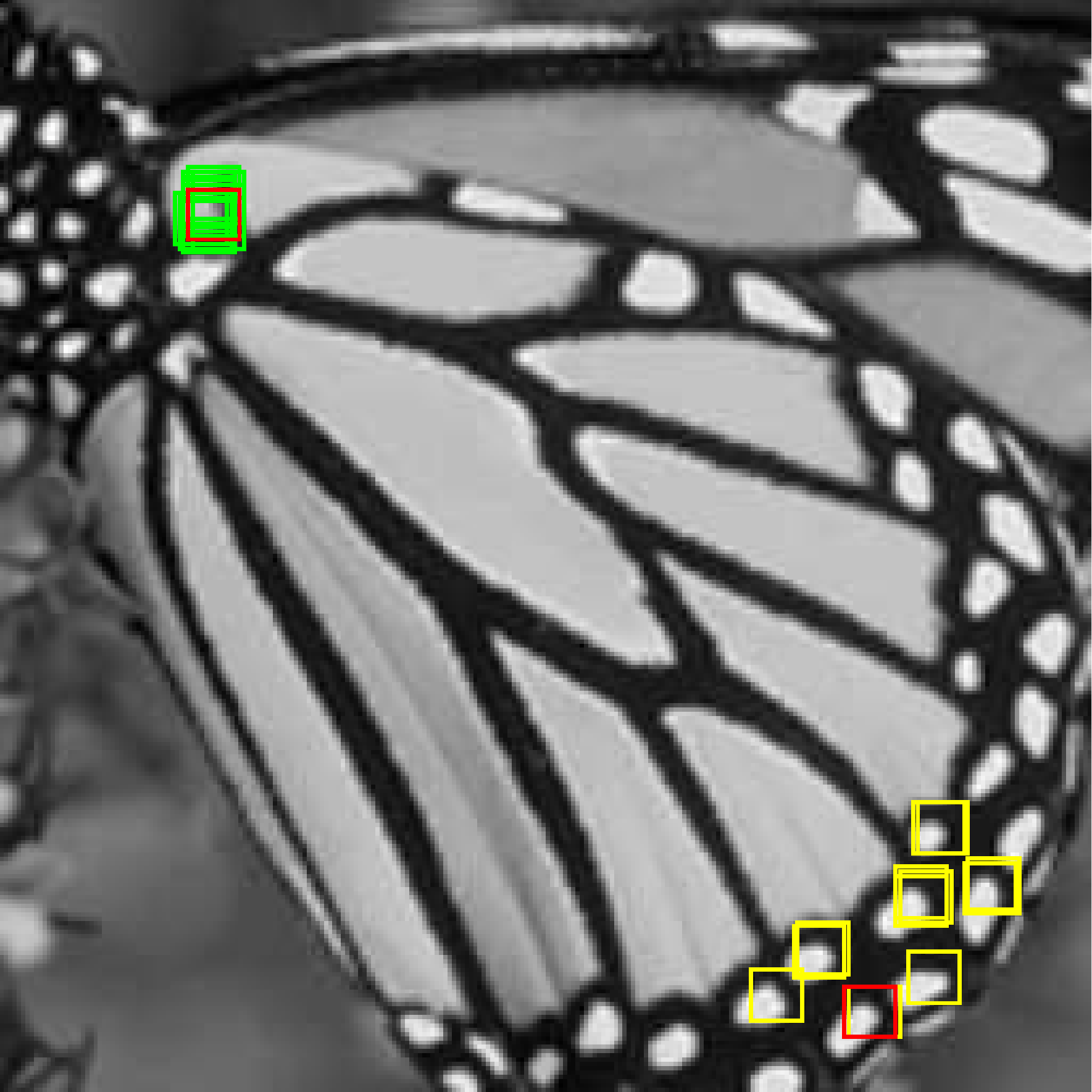}%
\label{fig:search:incremental}}

\caption{Three types of search strategies. Global, local and incremental. 
Red blocks indicate the reference patches. Green patches denote the matching 
patches for the reference patch at the top of the butterfly. Yellow patches 
denote the matching results corresponding to the reference at the bottom of the 
butterfly.}
\label{fig:search}
\end{figure*}

A straightforward solution to define the search window size for block matching
would be to use the whole image as the search space. In doing so, we would be in
the situation of global self-similarity and guarantee the selection of all the
available patches meeting the similarity constraint. There are, however, two
drawbacks to this solution. First, it incurs a significant computational
overhead as the complexity grows quadratically with the radius of the search
window. Second, it inevitably results in the inclusion of certain patches that,
although close to the reference patch in the Euclidean space, represent very
different structures in the image. This effect can be observed in
Fig.~\ref{fig:search:global}, specifically on the top patch, where global
self-similarity results in the selection of patches which do not lie on the
butterfly and have very different surrounding structure compared to the
reference patch. An alternative solution would be limit the search window to a
small neighborhood of the reference patch. However, if the search window is too
small, it might happen that not enough similar patches can be found, as
exemplified in Fig.~\ref{fig:search:local}. In our proposal we use an
incremental approach that starts with a small search window and enlarges it just
enough to find a full group of patches which exhibit an Euclidean distance to
the reference patch smaller that a preset
value. Fig.~\ref{fig:search:incremental} shows an example where this incremental
strategy finds similar patches from the local region for both reference blocks.

We tested the three different definitions of the search space here discussed,
aiming to find $32$ similar patches, resulting in Table~\ref{tab:search}. It can
can be observed that for some images, the use of global search results in a
drop of performance, while the use of incremental search never compromises the
reconstruction quality.

\begin{table}[t]
\centering
\caption{Search strategy effect on performance (\emph{Set5} x4). Run-time is a
factor relative to search strategy: global.}
\label{tab:search}
\input{table/search.tex}

\end{table}

\subsection{Iterative Procedure Parameters}
\label{sec:convergence}

\begin{figure}
	\centering
	\subfloat[PSNR plot]{%
		\pgfplotsset{width=.7\linewidth}%
		\input{plot/iteration_psnr_convergence.tex}%
	}
	\\
	\subfloat[Bird]{\includegraphics[width=.4\linewidth]{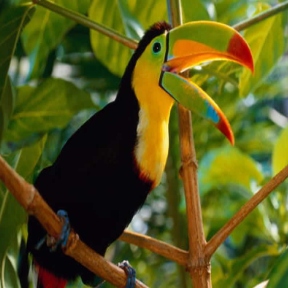}}
	\hspace{5mm}
	\subfloat[Butterfly]{\includegraphics[width=.4\linewidth]{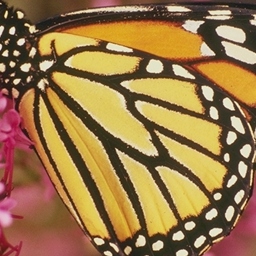}}

	\caption{Evolution of algorithm for different images. (a) The PSNR is
	computed for different number of iterations for two images (bird and
	butterfly) from \emph{Set5}. The algorithm progresses quickly for bird
	while for butterfly it requires more iterations. (b) Bird image has
	uniform regions and green textured regions, hence also evolves quickly.
	(c) Butterfly image has sharp edges which results in slow progression.}
	\label{fig:convergence}
\end{figure}

The iterative nature of the proposed solution introduces the need to select two
global parameters that significantly affect the overall system performance: the
total number of iterations and the collaborative filter strength curve,
$\tau_{\theta}$. We use an inverse square filter strength curve, with fixed
starting and end point, as described in the following equation:
\begin{equation}
	\tau_{\theta}^k = \gamma_k \frac{(K-k)}{K}^2 + \gamma_s s
	\label{equ:tau}
\end{equation}
Here $K$ is the total number of iterations, $k$ is the current iteration and
$s$ is the scale factor. This curve will lead to slower convergence when more
iterations are used and vice-versa, allowing the number of iterations to be
adjusted freely.

In order to devise a rule for the selection of the number of iterations, we
studied the convergence of the method by reconstruction various images of
\emph{Set5} using a different number of iterations. Figure~\ref{fig:convergence}
shows the results for the bird and butterfly images. These two images have a
very different type of content, and the reconstruction of the sharp edges
presented by the butterfly image requires a much slower variation of the
filtering strength, and therefore many more iterations, than the reconstruction
of the more smooth bird image. We speculate that this behavior stems from the
low pass nature of the employed sampling operator $H$ and devised a heuristic
that uses this known operator to compute the required number of iterations for a
particular image. This heuristic is presented together with other implementation
details in \ref{sec:parameters}, more specifically equation
(\ref{equ:iterations}).

\section{Experiments}
\label{sec:experiments}


\begin{figure*}[!ht]
\centering
\showdetailall{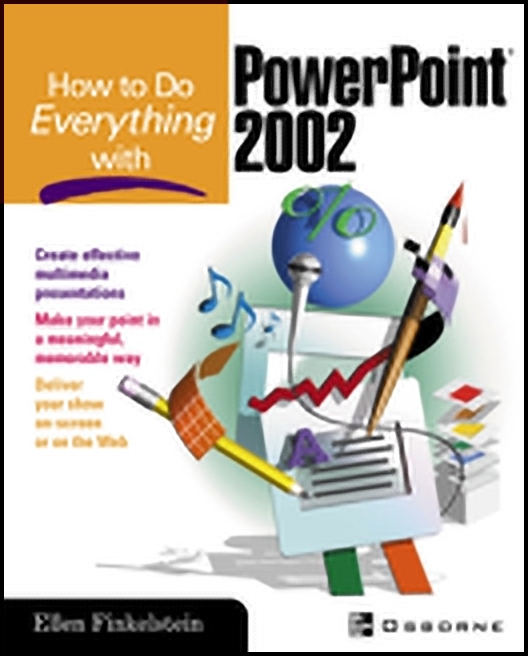}{.7cm,3cm}{3cm}{3cm}{3}{8cm}
\caption{Visual comparison with other approaches on the ppt image of 
\emph{Set14}, scale factor 4.}
\label{fig:ppt_analysis}
\end{figure*}

\begin{figure*}[!ht]
\centering
\showdetailall{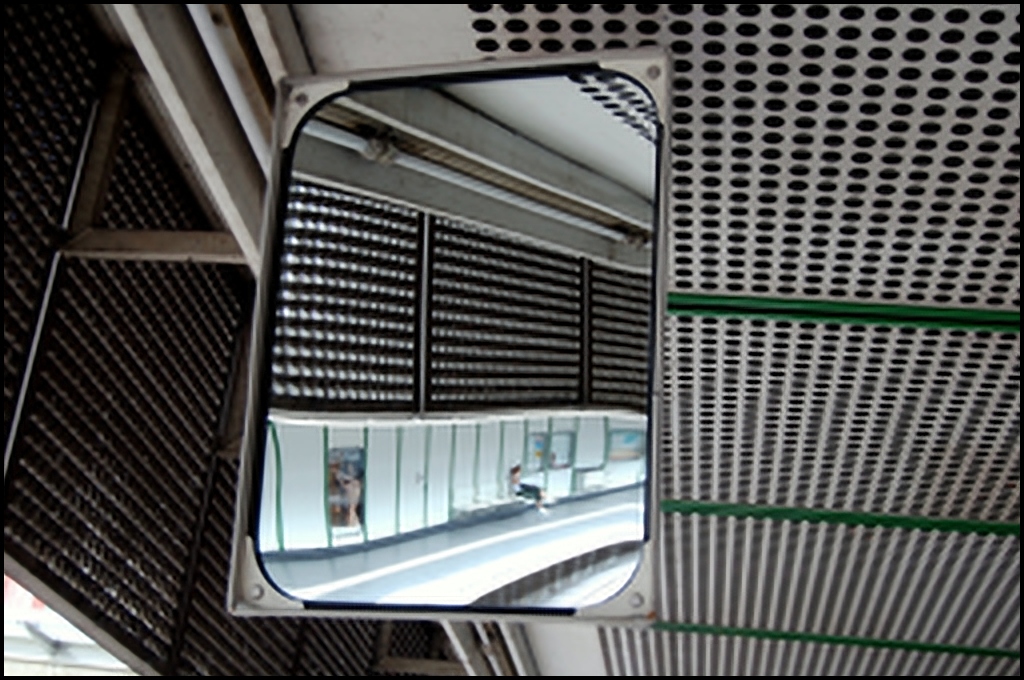}{3cm,1.2cm}{4cm}{2cm}{5}{6cm}
\caption{Visual comparison with other approaches on the 004 image of
\emph{Urban100}, scale factor 4.}
\label{fig:urban_analysis}
\end{figure*}

We evaluate the performance of the proposed \nameofthegame\ on three different
datasets and three scaling factors. First, we provide details on the datasets,
performance evaluation procedure and algorithm implementation. Next, the
selected parameters of the proposed method are presented. Then, the performance
of the proposed approach is compared with the state-of-the-art techniques, both
quantitatively and qualitatively.

All the experiments were conducted on a computer with an Intel Core
i7-4870HQ@2.5GHz, 16GB of RAM and an NVIDIA GeForce GT 750M. The \gls{wsdsr}
implementation used to generate these results can be accessed in the website:
\url{http://www.cs.tut.fi/sgn/imaging/sr/wsd/}.

\subsection{Experimental Setup}

\textbf{Datasets:} Following the recent work on \gls{sisr}, we test our approach
on three publicly available datasets. \emph{Set5} \cite{bevilacqua_2012_low} and
\emph{Set14} \cite{zeyde_2012_single} containing 5 and 14 images, respectively.
These two datasets have been extensively used by researchers to test
super-resolution algorithms, but are quite limited in both the amount and type
of images, containing mostly objects and people. For a more thorough analysis we
also test the proposed algorithm on the \emph{Urban100} dataset proposed by
\cite{huang_2015_single} which contains 100 images, including buildings and real
world structures. 



\textbf{Performance evaluation:} In order to evaluate the performance of the
proposed method we use a similar approach as \citet{timofte_2014_adjusted}. Color
images are converted to the YCbCr domain and only the luminance channel (Y) is
processed and evaluated. The color components, are taken into account for
display purposes alone, for which a bicubic interpolation is performed. The
evaluation of a method's performance using a scaling factor of $s$ on an image
$z_{orig}$, comprises the following steps:

\begin{enumerate}
	\item Set $z$ to the luminance channel of $z_{orig}$, which on color
	images corresponds to the Y component of the YCbCr color transform;
	\item Remove columns (on the right) and rows (on the bottom) from $z$ as
	needed to obtain an image which size is a multiple of $s$ on both width
	and height, designated $z_{gt}$;
	\item Quantize $z_{gt}$ using 8 bit resolution.
	\item Generate a low resolution image for processing by down-sampling
	$z_{gt}$ by a factor of $s$, using bicubic interpolation and an
	anti-aliasing filter, obtaining $z_{lr}$;
	\item Quantize $z_{lr}$ using 8 bit resolution;
	\item Super resolve $z_{lr}$, obtaining $y$;
	\item Quantize $y$ using 8 bit resolution;
	\item Remove a border of $s$ pixels from both $z_{gt}$ and $y$ obtaining
	$z_{gt\_trimmed}$ and $y_{trimmed}$;
	\item Compute the \gls{psnr} of $y_{trimmed}$ using as reference
	$z_{gt\_trimmed}$.
\end{enumerate}

We note that the trimming operations 2 and 8 are done in order to allow for fair
comparison with other methods. The proposed method can use any positive real
scaling factor and does not generate artefacts at the borders. The quantization
operations 3, 5 and 7 are used in order to effectively simulate a realistic
scenario where images are usually transmitted and displayed with 8 bit
resolution.

\subsection{Parameters}
\label{sec:parameters}

The \nameofthegame parameters, affecting both \nameofthefilter\ and the back
projection scheme used throughout these experiments are presented in
Table~\ref{tab:parameters}, where $s$ stands for the scale factor. Block size
$N_1$ is an important factor involved in the collaborative filtering which
depends on the up-sampling factor. The initial radius of the search window,
$N_S$, is set to $12$ for both steps. However, the \gls{ht} step uses
only local search, while the Wiener filter stage uses adaptive window size.
This difference is evident in the maximum search radius $N_{S_{max}}$. The
maximum number of used similar matches, $N_2$, is the same for both stages. The
regular grid used to select the reference blocks has step size, $N_{step}$,
defined such that there is $1$ pixel overlap between adjacent blocks. Finally,
the used transforms reflect the main goal of this work, that is, to perform the
Wiener filter only along the similarity domain. The total number of iterations
is computed using the following heuristic:

\begin{equation}
	K = \beta_1 * \frac{||y - HUy||^2_2}{m} + \beta_0
	\label{equ:iterations}
\end{equation}
where $U$ is the up-sampling operator matrix, and $m$ the dimensionality of $y$,
both as defined in section~\ref{sec:framework}. This will lead to the use of
more iterations in images that are more affected by the sampling procedure.
There is however an upper bound of $400$ iterations. 

\begin{table}
\centering
\caption{Proposed \nameofthegame\ parameters.}
\label{tab:parameters}
\begin{tabular}{l|ll}
\toprule
\multirow{7}{*}{HT stage parameters}
& $ N^{ht}_1 $      & $ max(8, 4*(s - 1)) $ \\
& $ N^{ht}_2 $      & $ 32 $ \\
& $ N^{ht}_{S_0} $     & $ 12 $ \\
& $ N^{ht}_{S_{max}} $ & $ 12 $ \\
& $ N^{ht}_{step} $ & $ N^{ht}_1 - 1 $ \\
& $ T^{ht}_{2D} $   & 2D-DCT  \\
& $ T^{ht}_{1D} $   & 1D-Haar \\\midrule
\multirow{7}{*}{Wiener stage parameters}
& $ N^{wiener}_1 $       & $ 0.5N^{ht}_1 $ \\
& $ N^{wiener}_2 $       & $ 32 $ \\
& $ N^{wiener}_{S_0} $     & $ 12 $ \\
& $ N^{wiener}_{S_{max}} $ & $ 48 $ \\
& $ N^{wiener}_{step} $  & $ N^{wiener}_1 - 1 $ \\
& $ T^{wiener}_{2D} $    & $I$ \\
& $ T^{wiener}_{1D} $    & 1D-Haar \\\midrule
\multirow{5}{*}{Global parameters}
& $ \alpha $   & $ 1.75 $\\
& $ \gamma_k $ & $ 12  $\\
& $ \gamma_s $ & $ 2/3 $\\
& $ \beta_1 $ & $ 40 / \sqrt{s} $\\
& $ \beta_0 $ & $ 20 $\\
& $ K^{pilot} $ & $ 5 $\\
\bottomrule

\end{tabular}
\end{table}

\subsection{Comparison with State-of-the-art}

The performance of the proposed approach is compared with several other methods
on the already mentioned datasets, using three up-sampling factors
$s = {2,3,4}$.  The results of the proposed approach are compared with the
classic bicubic interpolation, the regression based method \gls{aplus}
\cite{timofte_2014_adjusted}, the random forest based methods \gls{arflplus}
\cite{schulter_2015_fast} and \gls{nbsrf} \cite{salvador_2015_naive}, the
\gls{cnn} based methods \gls{vdsr} \cite{kim_2016_accurate} and \gls{drcn}
\cite{kim_2016_deeply} and finally the only self similarity based method on this
list, \gls{srex} \cite{huang_2015_single}. Furthermore, to highlight the
importance of using 1D Wiener in the second stage, we also present the
quantitative results achieved by our proposal when $T^{wiener}_{2D} = $ 2D-DCT,
designated \nameofthegame-DCT in Table~\ref{tab:performance_psnr}. \gls{psnr} is
used as the evaluation metric and the experimental procedure earlier explained
is used for all methods, with the notable exception of \gls{vdsr} and \gls{drcn}
for which the \gls{psnr} was computed on the publicly available results (only
steps 7 to 9 of the experimental procedure).

\subsubsection{Quantitative Analysis}

\begin{table*}[t]
	\centering
	\caption{The comparison of performance on \emph{Set5}, \emph{Set14} and
	\emph{Urban100}.}
	\label{tab:performance_psnr}
	\input{table/performance_psnr.tex}
\end{table*}

Table~\ref{tab:performance_psnr} shows the quantitative results of these
methods. It can be observed that the proposed approach outperforms all but the
more recent \gls{cnn} based methods: \gls{vdsr} and \gls{drcn}. Note that these
two methods used external data and reportedly require 4 hours and 6 days to
generate the necessary models, contrary to our approach that relies solely on
the image data. Comparing to the only other self-similarity based method,
\gls{srex} \cite{huang_2015_single}, the proposed method shows considerable
better performance, implying that the collaborative processing of the mutually
similar patches provides a much stronger prior than the single most similar
patch from the input image. We also note that for high up-sampling factors of
\emph{Urban100}, the performance of the proposed method is in par with even the
\gls{cnn} based methods, showing that this approach is especially suited for
images with a high number of edges and marked self-similarity. It also confirms
that hypothesis that the self-similarity based priors, although less in number,
are very powerful, and can compete with dictionaries learned over millions of
patches. Finally we note that the use of Wiener filter in similarity domain shows a significant performance improvement over the use of Wiener filter in 3D transform domain, which further supports our hypothesis that this specific feature is indeed crucial for the overall performance of the proposed approach.

\subsubsection{Qualitative Analysis}

So far we evaluated the proposed approach on a benchmark used for \gls{sisr}
performance assessment. Here we extend our analysis by providing a discussion on
the visual quality of the results obtained by various methods. The analysis is
conducted on results obtained with up-scaling factor of 4.

First we analyze a patch image ppt of \emph{Set14}, in
Fig.~\ref{fig:ppt_analysis}.  The background helps to notice the differences in
sharpness that results from the different techniques. It can be observed that
\nameofthegame\ estimates the high frequencies better than other approaches,
even the \gls{cnn} based ones, resulting in much sharper letters.

Finally, we consider a patch from image 004 of \emph{Urban100} in
Fig.~\ref{fig:urban_analysis}. The images in the Urban dataset exhibit a high
degree of self-similarity and the proposed approach works particularly well on
these kind of images. To illustrate,  we consider a patch which consists of
repetitive structure. It can be observed that the proposed approach yields much
sharper results than the others.

\subsection{Comparison with Varying Number of Iterations}
\label{sec:iterations}

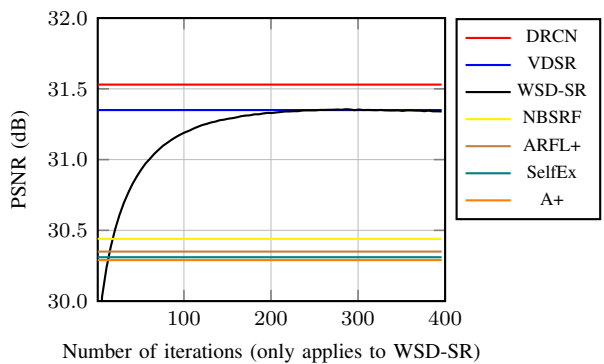
\begin{figure}
	\centering
	\pgfplotsset{width=.7\linewidth}
	\input{plot/iteration_psnr.tex}
	\caption{\label{fig:iteration_psnr} The average PSNR on \emph{Set5}, for
	a scale factor of 4. The variable number of iterations is only
	meaningful for \nameofthegame. All other methods were run with their
	\emph{canonical} configurations.}
\end{figure} 

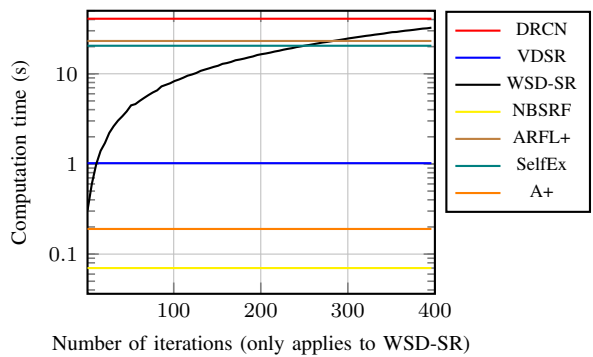
\begin{figure}
	\centering
	\pgfplotsset{width=.7\linewidth}
	\input{plot/iteration_time.tex}
	\caption{The average computation time on \emph{Set5}, for a scale
	factor of 4. The variable number of iterations is only meaningful for
	\nameofthegame. All other methods were run with their \emph{canonical}
	configurations.}
	\label{fig:iteration_time} 
\end{figure} 

We investigate the effect of having a fixed number of iterations on the
performance of the proposed approach, when compared with other approaches, as
opposed to using the estimation method presented in
Section~\ref{sec:convergence}.  Figure~\ref{fig:iteration_psnr} shows the
average \gls{psnr} on \emph{Set5} using an up-sampling factor of 4. We can see
that with a few dozen iterations our method outperforms most of the other
approaches, most notably the self-similarity based \gls{srex}. With a further
increase in number of iterations it is even capable of achieving similar results
as the state of the art convolutional network based approach \gls{vdsr}.

Next, we plot the computation time against the number of iterations in
Fig.~\ref{fig:iteration_time}. We also show the computation time of the other
approaches in a way that allows easy comparison. Note however that the number of
iterations is only relevant to \nameofthegame. All other approaches were 
executed in
their canonical state, using the publicly available codes.  As expected, for
\nameofthegame\ the computation time increases linearly with the number of 
iterations.
It can be observed that the proposed approach is generally slower than the
dictionary based methods. Note also that even at 400 iterations, the proposed
approach still performs faster than the only method for which we can't match the
reconstruction performance, \gls{drcn}. Compared to the self-similarity based
approach \cite{huang_2015_single}, the proposed algorithm is able to achieve
comparable results much faster, and about 1dB better at the break even point. In
\nameofthegame, the number of iterations can provide a trade-off between the
performance and the processing time of the algorithm.

\section{Discussion}
\label{sec:discussion}

Here we study a few variations of \nameofthegame. First we propose and evaluate 
it's
extension to color images. Second, we analyze the method's
performance under the assumption that a block match oracle is available, in
order to assess the existence of potential for better results.

\subsection{Color Image Channels}

Following the established custom, all the tests and comparisons so far have been
conducted using only the luminance information from the input images. Despite
the fact that this channel contains most of the relevant information, we believe
that some gain might come from making use of the Color channels in the
reconstruction process. Our method is easily extended to such scenarios, and we
devised and tested two new profiles in order to verify this hypothesis. The
first profile, termed Y-YCbCr follows a similar approach as presented in
\cite{dabov_2007_color}, where the block matching is performed in the Y channel
and the filtering applied to all the Y, Cb and Cr channels. The second profile,
Y-RGB also performs the block matching in the Y channel, but does the filtering
on all channels of the RGB domain. We show in Table~\ref{tab:icf_colour} the 
results of
processing \emph{Set5} with these profiles. We add a third profile in the table,
named Y-Y that corresponds to the one we have been using so far that uses only
information from the Y channel for both matching and filtering and that super
resolves the chrominance channels with a simple bicubic interpolator.

\begin{table}[ht]
	\centering
	\caption{\nameofthegame\ Performance When Color Information is Used}
	\label{tab:icf_colour}
	\input{table/icf_colour.tex}

\end{table}

Despite the fact that the Y-RGB profile does not filter the Y channel, it is
possible to see that, even when measured as the \gls{psnr} of the Y channel
alone, the method's performance improves considerably on the bird and butterfly
images.


\subsection{Oracles}

We performed a final experiment which we believe shows the potential of this
technique to achieve even better results. This experiment was conducted with the
use of oracles, more specifically an oracle for the block match table. This
match table was extracted from the ground truth and used on all iterations,
while all other parameters of the method remained as previously defined. In
essence we are substituting the block matching procedure with an external
entity, the oracle, that provides the best possible match table. The results
from this experiment, conducted on \emph{Set5} using a scale factor of 4, can
be observed in Table~\ref{tab:icf_oracle_match}. As one can see, also here, 
there is
potential for much better results if the block matching procedure is somehow
improved.

\begin{table}[h]
	\centering
	\caption{\nameofthegame\ Performance when Oracle Provides Matches}
	\label{tab:icf_oracle_match}
	\input{table/icf_oracle_match.tex}

\end{table}

%

%

\section{Conclusion}
\label{sec:conclusion}

Our previous algorithm employing iterative back-projection for \gls{sisr}
\citep{egiazarian_2015_single} made use of a collaborative filter designed for
denoising applications, \gls{bm3d}, which uses a 3D Wiener filter in groups of
similar patches. In this work, we have shown that 1D Wiener filtering along the
similarity domain is more effective for the specific problem of \gls{sisr} and
results in much sharper reconstructions. Our novel collaborative filter,
\gls{wsd}, is able to achieve state-of-the-art results when coupled with
iterative back-projection, a combination we termed \gls{wsdsr}. Furthermore, the
use of self-similarity prior leads to a solution that does not need training and
relies only on the input image.

The summary of our findings is:
\begin{itemize}
\item 1D Wiener filtering along similarity domain is more effective than 3D
Wiener filtering for the task of \gls{sisr};
\item Local self-similarity produces more relevant patches than global
self-similarity; 
\item The patches extracted from input image can provide strong prior for
\gls{sisr}.
\end{itemize}

We demonstrated empirically that the proposed approach works well not only on
images with substantial self-similarity but also on natural images with more
complex textures. We have also shown that there is still potential within this
framework, more specifically, the performance can be improved by: (1) taking
advantage of the color information and (2) improving the block matching
strategy.

\bibliographystyle{IEEEtranSN}
\bibliography{IEEEabrv,bibliography}









\begin{IEEEbiography}[{\includegraphics[width=1in,height=1.25in,clip,keepaspectratio]{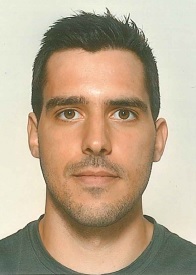}}]{Cristóvão Cruz}
Cristóvão Cruz completed his masters degree in Electronic and
Telecommunications Engineering at University of Aveiro in 2014. He is currently
working at Noiseless Imaging Oy (Ltd) as an algorithms engineer and is enrolled on
the doctoral programme of Computing and Electrical Engineering at Tampere
University of Technology. His current research is focused on the design and
implementation of image restoration algorithms.
\end{IEEEbiography}

\begin{IEEEbiography}[{\includegraphics[width=1in,height=1.25in,clip,keepaspectratio]{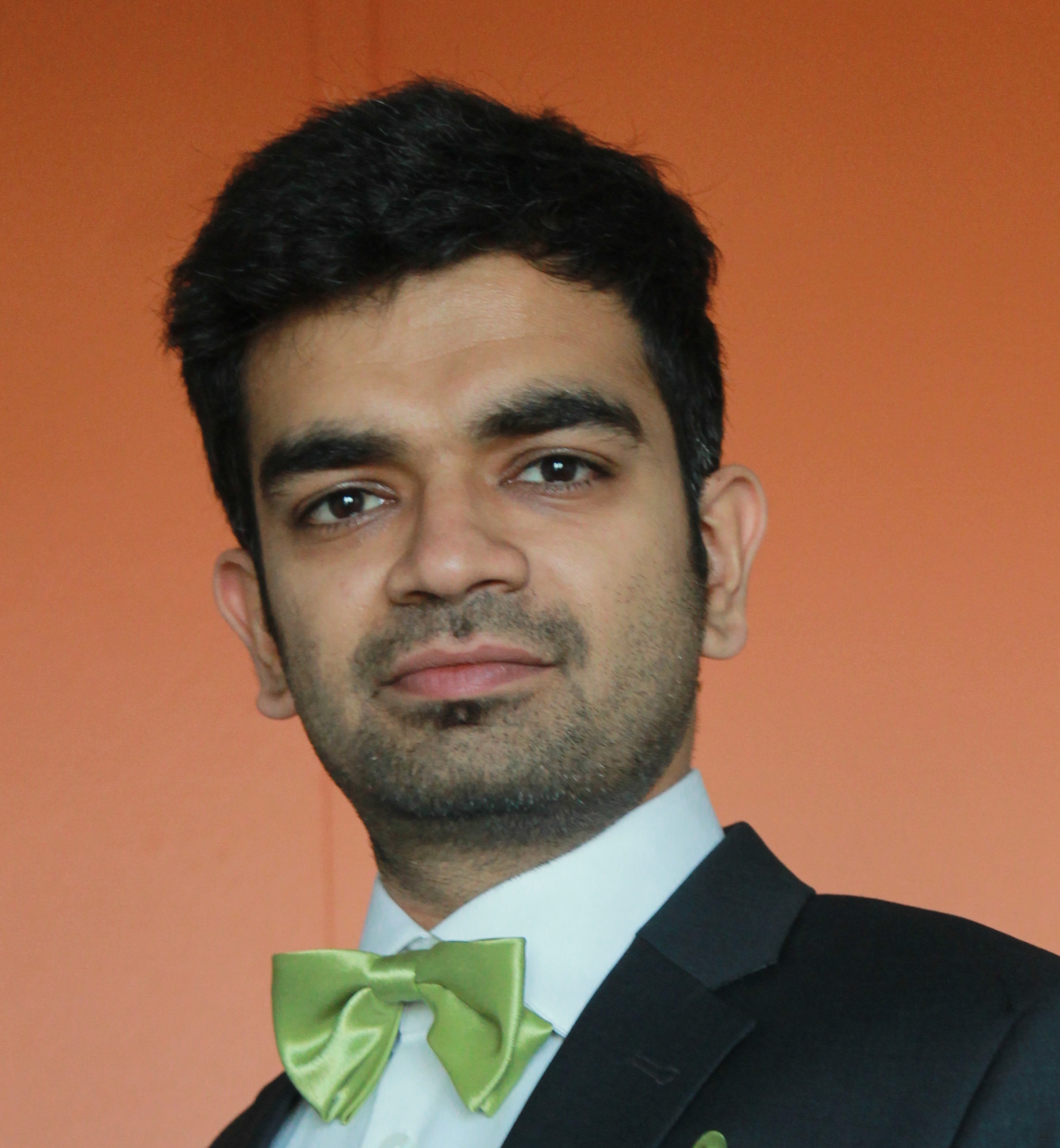}}]{Rakesh Mehta}
Rakesh Mehta received the M.S. degree in 2011, and Ph.D. degree in 2016 from Tampere University of Technology, Finland. He is currently working as a senior research scientist at United Technologies Research Centre In Cork, Ireland. His research interests include object detection, feature description, texture classification and deep learning.
\end{IEEEbiography}

\begin{IEEEbiography}[{\includegraphics[width=1in,height=1.25in,clip,keepaspectratio]{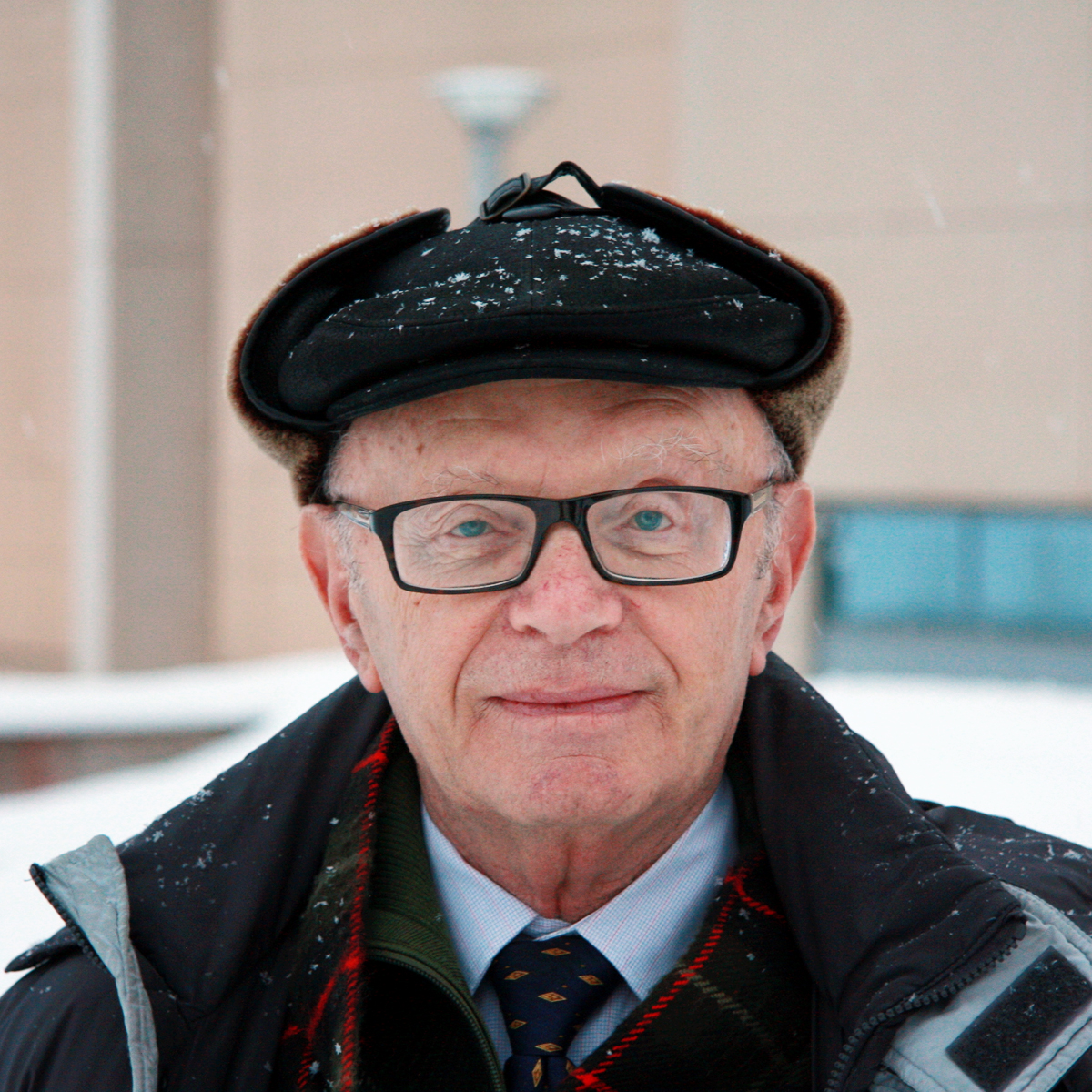}}]{Vladimir Katkovnik}
Vladimir Katkovnik received Ph.D. and D.Sc. degrees in technical cybernetics from Leningrad Polytechnic Institute (LPI), Leningrad, Russia, in 1964 and 1974, respectively. From 1964 to 1991, he was an Associate Professor and then a Professor with the Department of Mechanics and Control Processes, LPI. From 1991 to 1999, he was a Professor with the Department of Statistics, University of South Africa, Pretoria, Republic of South Africa. From 2001 to 2003, he was a Professor with Kwangju Institute of Science and Technology, Gwangju, South Korea. Since 2003, he is with the Department of Signal Processing, Tampere University of Technology, Tampere, Finland. He published 6 books and over 300 papers. His research interests include stochastic image/signal processing, linear and nonlinear filtering, nonparametric estimation, computational imaging, nonstationary systems, and time–frequency analysis.
\end{IEEEbiography}

\begin{IEEEbiography}[{\includegraphics[width=1in,height=1.25in,clip,keepaspectratio]{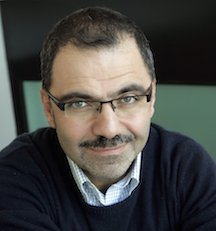}}]{Karen Egiazarian}
Karen O. Egiazarian (Eguiazarian) (Senior Member IEEE, 1996) received M.Sc. in mathematics from Yerevan State University, Armenia, in 1981, the Ph.D. degree in physics and mathematics from Moscow State University, Russia, in 1986, and a Doctor of Technology from Tampere University of Technology, Finland, in 1994. 
In 2015 he has received the Honorary Doctoral degree from Don State-Technical University (Rostov-Don, Russia).  
Dr. Egiazarian is a co-founder and CEO of Noiseless Imaging Oy (Ltd), Tampere University of Technology spin-off company.
He is a Professor at Signal Processing Laboratory, Tampere University of Technology, Tampere, Finland, leading ‘Computational imaging’ group, head of Signal Processing Research Community (SPRC) at TUT, and Docent in the Department of Information Technology, University of Jyväskyla, Finland. His main research interests are in the field of computational imaging, compressed sensing, efficient signal processing algorithms, image/video restoration and compression. Dr. Egiazarian has published about 700 refereed journal and conference articles, books and patents in these fields. 
Dr. Egiazarian has served as Associate Editor in major journals in the field of his expertise, including IEEE Transactions on Image Processing. He is currently an Editor-in-Chief of Journal of Electronic Imaging (SPIE), and Member of the DSP Technical Committee of the IEEE Circuits and Systems Society.
\end{IEEEbiography}

\vfill

\end{document}

%% file: acronyms.tex
\newacronym{aplus}{A+}{A+}
\newacronym{arflplus}{ARFL+}{ARFL+}
\newacronym{ann}{ANN}{approximate nearest neighbours}
\newacronym{bm}{BM}{Block Matching}
\newacronym{bm3d}{BM3D}{Block Matching 3D}
\newacronym{cda}{CDA}{coupled deep autoencoder}
\newacronym{cnn}{CNN}{convolutional neural network}
\newacronym{gpu}{GPU}{graphical processing units}
\newacronym{fri}{FRI}{finite rate of innovation}
\newacronym{hr}{HR}{high resolution}
\newacronym{ht}{HT}{hard-thresholding}
\newacronym{lle}{LLE}{locally linear embedding}
\newacronym{lr}{LR}{low resolution}
\newacronym{nbsrf}{NBSRF}{NBSRF}
\newacronym{psnr}{PSNR}{peak signal to noise ratio}
\newacronym{sisr}{SISR}{single image super-resolution}
\newacronym{srcnn}{SRCNN}{SRCNN}
\newacronym{srex}{SelfEx}{SelfEx}
\newacronym{ia}{IA}{IA}
\newacronym{vdsr}{VDSR}{VDSR}
\newacronym{drcn}{DRCN}{DRCN}
\newacronym{wsd}{WSD}{Wiener filter in Similarity Domain}
\newacronym{wsdsr}{WSD-SR}{Wiener filter in Similarity Domain for Super Resolution}

%% file: figure/flowchart.tikz

\def\blkfnt{\scriptsize}

\begin{tikzpicture}[
	op/.style={node distance=4mm,align=center,draw},
	image/.style={inner sep=0},
	inner sep=2mm,
]
	\node[image] (basic input) {\includegraphics[width=2cm]{set5_002}};
	\node[op,right=of basic input] (basic block match) {\blkfnt BLOCK\\\blkfnt 
MATCH};

	\node[op,right=1 cm of basic block match] (basic group) {\blkfnt GROUP};
	\node[op,right=of basic group] (basic transform)  {$T_{2D}$\\$+$\\$T_{1D}$};
	\node[op,right=of basic transform] (basic shrink) {\blkfnt HT};
	\node[op,right=of basic shrink] (basic inverse) 
{$T^{-1}_{2D}$\\$+$\\$T^{-1}_{1D}$};
	\node[op,right=of basic inverse] (basic aggregation) {\blkfnt AGGREGATE};
	
	\node[image,below=1.3cm of basic aggregation,anchor=north] (pilot)
{\includegraphics[width=2cm]{set5_002}};
	\node[below=.5mm of pilot] (pilot label){\scriptsize PILOT($\tilde{x}^{pilot}$)};

	\node[op,left=of pilot] (pilot block match) {\blkfnt BLOCK\\\blkfnt MATCH};
	\node[op,left=of pilot block match] (pilot group) {\blkfnt GROUP};
	\node[op,left=of pilot group] (pilot transform) {$T_{1D}$};
	
	\node[op,below=1.3cm of pilot] (final group) {\blkfnt GROUP};
	\node[op,left=of final group] (final transform) {$T_{1D}$};
	
	\node[op,left=of final transform] (wiener) {\blkfnt WIENER};

	\node[op,left=of wiener] (final inverse)  {$T^{-1}_{1D}$};
	\node[op,left=of final inverse] (final aggregation) {\blkfnt AGGREGATE};

	\node[image,left=1cm of final aggregation] (final)
{\includegraphics[width=2cm]{set5_002}};

	\coordinate[above=8mm of basic group.north west] (basic group above);
	\coordinate[left=1cm of basic group above] (basic match above);
	\coordinate[below=8mm of basic block match.south west] (basic match below);
	
	\node[draw,dashed,fit=(basic match above) (basic match below)] (basic match) 
{};
	\node[inner sep=1mm] at (basic match.north west) [above right] 
{\scriptsize{$k = 0$}};
	
	\node[draw,dashed,fit=(pilot) (basic transform) (basic group above) (basic 
aggregation) (pilot label)] (pilot stuff) {};
	\node[inner sep=1mm] at (pilot stuff.north east) [above left] 
{\scriptsize{$k \mod K^{pilot} = 0$}};

	\draw[->] (basic input) -- (basic block match);
	\draw[->] (basic block match) -- (basic group);
	\draw[->] (basic group) -- (basic transform);
	\draw[->] (basic transform) -- (basic shrink);
	\draw[->] (basic shrink) -- (basic inverse);
	\draw[->] (basic inverse) -- (basic aggregation);
	\draw[->] (basic aggregation) -- (pilot);
	\draw[->] (pilot) -- (pilot block match);
	\draw[->] (pilot block match) -- (pilot group);
	\draw[->] (pilot group) -- (pilot transform);
	\draw[->] (pilot transform) -- (wiener);
	\draw[->] (pilot block match) -- (final group);
	\draw[->] (final group) -- (final transform);
	\draw[->] (final transform) -- (wiener);
	\draw[->] (wiener) -- (final inverse);
	\draw[->] (final inverse) -- (final aggregation);
	\draw[->] (final aggregation) -- (final);
	
	\coordinate[above=1cm of basic input] (basic input above);
	\coordinate[right=1.5cm of final group] (final group right);
	\draw[->] (basic input) -- (basic input above) -| (final group right) -- 
(final group);
	\draw[->] (basic input above) -| (basic group);
	
	\coordinate[above left=3mm of pilot] (pilot above left);
	
	\draw[->] (pilot) -- (pilot above left) -| (pilot group);
	
	\node[below=.5mm of final] {\scriptsize OUTPUT ($\tilde{x}^{k+1}$)};
	\node[below=.5mm of basic input] {\scriptsize INPUT ($x^k$)};
\end{tikzpicture} 

%% file: table/transform.tex
\pgfplotstableread{data/oned_transform_dct_skip.csv}\dctskip
\pgfplotstableread{data/oned_transform_normal.csv}\normal

\pgfplotstablenew[
	create on use/names/.style={
		create col/set list={Baby,Bird,Butterfly,Head,Woman,Average},
	},
	create on use/psnr2/.style={
		create col/copy column from table={\dctskip}{psnr},
	},
	create on use/psnr3/.style={
		create col/copy column from table={\normal}{psnr},
	},
	create on use/time2/.style={
		create col/copy column from table={\dctskip}{time},
	},
	columns={names,psnr2,psnr3,time2},
]{6}\thetable

\pgfplotstabletypeset[
	columns/names/.style={
		string type,
		column type={l|},
		column name={},
	},
	columns/psnr2/.style={column name=PSNR,column type=c|},
	columns/psnr3/.style={column name=PSNR},
	columns/time2/.style={column name=Speedup},
	every head row/.style={
		before row={%
			\toprule%
			$T^{2D}_{Wiener}$&
			DCT &
			\multicolumn{2}{c}{Identity*}\\\midrule
		},
		after row=\midrule,
	},
	every nth row={5}{before row=\midrule},
]{\thetable}

%% file: table/skip.tex
\pgfplotstableread{data/block_match_skip_block_match_skip_disabled.csv}\noskip
\pgfplotstableread{data/block_match_skip_normal.csv}\normal

\pgfplotstablenew[
	create on use/names/.style={
		create col/set list={Baby,Bird,Butterfly,Head,Woman,Average},
	},
	create on use/psnr1/.style={
		create col/copy column from table={\noskip}{psnr},
	},
	create on use/psnr2/.style={
		create col/copy column from table={\normal}{psnr},
	},
	create on use/time1/.style={
		create col/copy column from table={\noskip}{time},
	},
	columns={names,psnr1,psnr2,time1},
]{6}\thetable

\pgfplotstabletypeset[
	columns/names/.style={
		string type,
		column type={l|},
		column name={},
	},
	columns/psnr1/.style={column name=PSNR,column type=c|},
	columns/psnr2/.style={column name=PSNR},
	columns/time1/.style={column name=Speedup},
	every head row/.style={
		before row={%
			\toprule%
			Match table reuse &
			Disabled &
            		\multicolumn{2}{c}{Enabled*}\\\midrule
		},
		after row=\midrule,
	},
	every nth row={5}{before row=\midrule},
]{\thetable}

%% file: table/search.tex
%
%
%
\pgfplotstableread{data/search_strategy_local.csv}\local
\pgfplotstableread{data/search_strategy_global.csv}\ssglobal
\pgfplotstableread{data/search_strategy_normal.csv}\normal

\pgfplotstablenew[
	create on use/names/.style={
		create col/set list={Baby,Bird,Butterfly,Head,Woman,Average},
	},
	create on use/psnr1/.style={
		create col/copy column from table={\local}{psnr},
	},
	create on use/psnr2/.style={
		create col/copy column from table={\ssglobal}{psnr},
	},
	create on use/psnr3/.style={
		create col/copy column from table={\normal}{psnr},
	},
	create on use/time1/.style={
		create col/copy column from table={\local}{time},
	},
	create on use/time2/.style={
		create col/copy column from table={\ssglobal}{time},
	},
	columns={names,psnr2,psnr1,time1,psnr3,time2},
]{6}\thetable

\pgfplotstabletypeset[
	columns/names/.style={
		string type,
		column type={l|},
		column name={},
	},
	columns/psnr1/.style={column name=PSNR},
	columns/psnr2/.style={column name=PSNR, column type=c|},
	columns/psnr3/.style={column name=PSNR},
	columns/time1/.style={column name=Speedup, column type=c|},
	columns/time2/.style={column name=Speedup},
	every head row/.style={
		before row={%
			\toprule%
			Search Strategy &
			Global &
			\multicolumn{2}{c|}{Local} &
			\multicolumn{2}{c}{Incremental*}\\\midrule
		},
		after row=\midrule,
	},
	every nth row={5}{before row=\midrule},
]{\thetable}

%% file: plot/iteration_psnr_convergence.tex
\begin{tikzpicture}
\begin{axis}[
	xlabel=Number of iterations,
	ylabel=PSNR (dB),
	table/x=iterations,
	no markers,
	xmin=1,
	xmax=400,
	ymin=20,
	ymax=35,
	legend style={
		at={(0.97,0.03)},
		anchor=south east
	},
	grid=major,
	xticklabel style={
		/pgf/number format/precision=0,
	},
	yticklabel style={
		/pgf/number format/precision=0,
	},
]
	\pgfplotstableread{data/iteration_psnr_convergence.csv}\data
	\addplot table [y=bird] {\data};
	\addplot table [y=butterfly] {\data};

	\legend{Bird, Butterfly}
\end{axis}
\end{tikzpicture}

%% file: table/performance_psnr.tex

\pgfplotstableread{data/performance_psnr_set5.csv}\setfive
\pgfplotstableread{data/performance_psnr_set14.csv}\setfourteen
\pgfplotstableread{data/performance_psnr_urban100.csv}\urbanhundred

\pgfplotstablevertcat{\fulltable}{\setfive}
\pgfplotstablevertcat{\fulltable}{\setfourteen}
\pgfplotstablevertcat{\fulltable}{\urbanhundred}

\pgfplotstabletypeset[
	create on use/dataset/.style={create col/set list={Set5,,,Set14,,,Urban100,,}},
	columns/dataset/.style={
		string type,
		column name=Dataset,
		assign cell content/.code={
			\pgfmathparse{int(Mod(\pgfplotstablerow, 3))}%
			\ifnum\pgfmathresult=0%
				\pgfkeyssetvalue{/pgfplots/table/@cell content}
				{\multirow{3}{*}{\emph{##1}}}%
			\else
				\pgfkeyssetvalue{/pgfplots/table/@cell content}{}%
			\fi
		},
	},
	every nth row={3}{before row=\midrule},
	columns/factor/.style={column name=Factor, precision=0},
	columns/srex/.style={column name=SelfEx},
	columns/aplus/.style={column name=A+},
	columns/nbsrf/.style={column name=NBSRF},
	columns/arflplus/.style={column name=ARFL+},
	columns/wsdsr/.style={column name=\nameofthegame},
	columns/wsdsr_dct/.style={column name=\nameofthegame-DCT},
	columns/bicubic/.style={column name=Bicubic},
	columns/vdsr/.style={column name=VDSR},
	columns/drcn/.style={column name=DRCN},
	columns={dataset, factor, bicubic, aplus, srex, arflplus, nbsrf, vdsr, drcn, wsdsr_dct, wsdsr},
]{\fulltable}

%% file: plot/iteration_psnr.tex
\begin{tikzpicture}
\begin{axis}[
	xlabel=Number of iterations (only applies to \nameofthegame),
	ylabel=PSNR (dB),
	table/x=iterations,
	no markers,
	xmin=1,
	xmax=400,
	ymin=30,
	ymax=32,
	legend style={
		at={(1.03,1.00)},
		anchor=north west
	},
	grid=major,
	xticklabel style={
		/pgf/number format/precision=0,
	},
	yticklabel style={
		/pgf/number format/precision=1,
	},
]
	\pgfplotstableread{data/iteration_psnr.csv}\data
	\addplot table [y=drcn] {\data};
	\addplot table [y=vdsr] {\data};
	\addplot table [y=icf] {\data};
	\addplot table [y=nbsrf] {\data};
	\addplot table [y=arflplus] {\data};
	\addplot table [y=srex] {\data};
	\addplot table [y=aplus] {\data};

	\legend{DRCN, VDSR, \nameofthegame, NBSRF, ARFL+, SelfEx, A+}
\end{axis}
\end{tikzpicture}

%% file: plot/iteration_time.tex
\begin{tikzpicture}
\begin{semilogyaxis}[
	xlabel=Number of iterations (only applies to \nameofthegame),
	ylabel=Computation time (s),
	table/x=iterations,
	no markers,
	xmin=1,
	xmax=400,
	ymax=50,
	legend style={
		at={(1.03,1.00)},
		anchor=north west
	},
	grid=major,
	xticklabel style={
		/pgf/number format/precision=0,
	},
	log ticks with fixed point,
]
	\pgfplotstableread{data/iteration_time.csv}\data
	\addplot table [y=drcn] {\data};
	\addplot table [y=vdsr] {\data};
	\addplot table [y=icf] {\data};
	\addplot table [y=nbsrf] {\data};
	\addplot table [y=arflplus] {\data};
	\addplot table [y=srex] {\data};
	\addplot table [y=aplus] {\data};

	\legend{DRCN, VDSR, \nameofthegame, NBSRF, ARFL+, SelfEx, A+}
\end{semilogyaxis}
\end{tikzpicture}

%% file: table/icf_colour.tex
\pgfplotstableread{data/icf_colour.csv}\thetable

\pgfplotstabletypeset[
	create on use/names/.style={
		create col/set list={Baby,Bird,Butterfly,Head,Woman,Average},
	},
	columns/names/.style={
		string type,
		column type={l},
		column name={},
	},
	columns/normal_psnr_y/.style={column name=Y-Y},
	columns/colour_ycbcr_psnr_y/.style={column name=Y-YCbCr},
	columns/colour_rgb_psnr_y/.style={column name=Y-RGB},
	every nth row={5}{before row=\midrule},
	columns={
		names,
		normal_psnr_y,
		colour_ycbcr_psnr_y,
		colour_rgb_psnr_y
	},
]{\thetable}

%% file: table/icf_oracle_match.tex
\pgfplotstableread{data/icf_oracle_match_hand.csv}\thetable

\pgfplotstabletypeset[
	create on use/names/.style={
		create col/set list={Baby,Bird,Butterfly,Head,Woman,Average},
	},
	columns/names/.style={
		string type,
		column type={l},
		column name={},
	},
	columns/normal_psnr_y/.style={column name=Without Oracle},
	columns/oracle_match_both_psnr_y/.style={column name=With Oracle},
	every nth row={5}{before row=\midrule},
	columns={
		names,
		normal_psnr_y,
		oracle_match_both_psnr_y
	},
]{\thetable}

%% file: document.bbl
\begin{thebibliography}{49}
\providecommand{\natexlab}[1]{#1}
\providecommand{\url}[1]{#1}
\csname url@samestyle\endcsname
\providecommand{\newblock}{\relax}
\providecommand{\bibinfo}[2]{#2}
\providecommand{\BIBentrySTDinterwordspacing}{\spaceskip=0pt\relax}
\providecommand{\BIBentryALTinterwordstretchfactor}{4}
\providecommand{\BIBentryALTinterwordspacing}{\spaceskip=\fontdimen2\font plus
\BIBentryALTinterwordstretchfactor\fontdimen3\font minus
  \fontdimen4\font\relax}
\providecommand{\BIBforeignlanguage}[2]{{%
\expandafter\ifx\csname l@#1\endcsname\relax
\typeout{** WARNING: IEEEtranSN.bst: No hyphenation pattern has been}%
\typeout{** loaded for the language `#1'. Using the pattern for}%
\typeout{** the default language instead.}%
\else
\language=\csname l@#1\endcsname
\fi
#2}}
\providecommand{\BIBdecl}{\relax}
\BIBdecl

\bibitem[Ahmed and Shah(2016)]{ahmed_2016_single}
J.~Ahmed and M.~A. Shah, ``Single image super-resolution by directionally
  structured coupled dictionary learning,'' \emph{{EURASIP} Journal on Image
  and Video Processing}, vol.~36, no.~1, 2016.

\bibitem[Bevilacqua et~al.(2012)Bevilacqua, Roumy, Guillemot, and line
  Alberi~Morel]{bevilacqua_2012_low}
M.~Bevilacqua, A.~Roumy, C.~Guillemot, and M.~line Alberi~Morel,
  ``Low-complexity single-image super-resolution based on nonnegative neighbor
  embedding,'' in \emph{Proceedings of the British Machine Vision
  Conference}.\hskip 1em plus 0.5em minus 0.4em\relax BMVA Press, 2012, pp.
  135.1--135.10.

\bibitem[Buades et~al.(2005)Buades, Coll, and Morel]{buades_2005_non}
A.~Buades, B.~Coll, and J.-M. Morel, ``A non-local algorithm for image
  denoising,'' in \emph{IEEE Conference on Computer Vision and Pattern
  Recognition}, vol.~2, 2005, pp. 60--65.

\bibitem[Chang et~al.(2004)Chang, Yeung, and Xiong]{chang_2004_super}
H.~Chang, D.-Y. Yeung, and Y.~Xiong, ``Super-resolution through neighbor
  embedding,'' in \emph{IEEE Conference on Computer Vision and Pattern
  Recognition}, vol.~1, 2004, pp. I--I.

\bibitem[Choi and Kim(2017)]{choi_2017_single}
J.-S. Choi and M.~Kim, ``Single image super-resolution using global regression
  based on multiple local linear mappings,'' \emph{{IEEE} Transactions on Image
  Processing}, pp. 1--1, 2017.

\bibitem[Cui et~al.(2014)Cui, Chang, Shan, Zhong, and Chen]{cui_2014_deep}
Z.~Cui, H.~Chang, S.~Shan, B.~Zhong, and X.~Chen, ``Deep network cascade for
  image super-resolution,'' in \emph{Computer Vision--ECCV 2014}.\hskip 1em
  plus 0.5em minus 0.4em\relax Springer, 2014, pp. 49--64.

\bibitem[Dabov et~al.(2007{\natexlab{a}})Dabov, Foi, Katkovnik, and
  Egiazarian]{dabov_2007_color}
K.~Dabov, A.~Foi, V.~Katkovnik, and K.~Egiazarian, ``Color image denoising via
  sparse 3d collaborative filtering with grouping constraint in
  luminance-chrominance space,'' in \emph{2007 IEEE International Conference on
  Image Processing}, vol.~1, Sept 2007, pp. I -- 313--I -- 316.

\bibitem[Dabov et~al.(2007{\natexlab{b}})Dabov, Foi, Katkovnik, and
  Egiazarian]{dabov_2007_image}
------, ``Image denoising by sparse {3-D} transform-domain collaborative
  filtering,'' \emph{IEEE Transactions on Image Processing}, vol.~16, no.~8,
  pp. 2080--2095, 2007.

\bibitem[Danielyan et~al.(2012)Danielyan, Katkovnik, and
  Egiazarian]{danielyan_2012_bm3d}
A.~Danielyan, V.~Katkovnik, and K.~Egiazarian, ``Bm3d frames and variational
  image deblurring,'' \emph{IEEE Transactions on Image Processing}, vol.~21,
  no.~4, pp. 1715--1728, April 2012.

\bibitem[Deng et~al.(2016)Deng, Guo, and Huang]{deng_2016_single}
L.-J. Deng, W.~Guo, and T.-Z. Huang, ``Single-image super-resolution via an
  iterative reproducing kernel hilbert space method,'' \emph{{IEEE}
  Transactions on Circuits and Systems for Video Technology}, vol.~26, no.~11,
  pp. 2001--2014, nov 2016.

\bibitem[Dong et~al.(2014)Dong, Loy, He, and Tang]{dong_2014_learning}
C.~Dong, C.~C. Loy, K.~He, and X.~Tang, ``Learning a deep convolutional network
  for image super-resolution,'' in \emph{Computer Vision--ECCV 2014}.\hskip 1em
  plus 0.5em minus 0.4em\relax Springer, 2014, pp. 184--199.

\bibitem[Dong et~al.(2016)Dong, Loy, He, and Tang]{dong_2016_image}
------, ``Image super-resolution using deep convolutional networks,''
  \emph{{IEEE} Transactions on Pattern Analysis and Machine Intelligence},
  vol.~38, no.~2, pp. 295--307, feb 2016.

\bibitem[Dong et~al.(2013)Dong, Zhang, Lukac, and Shi]{dong_2013_sparse}
W.~Dong, L.~Zhang, R.~Lukac, and G.~Shi, ``Sparse representation based image
  interpolation with nonlocal autoregressive modeling,'' \emph{IEEE
  Transactions on Image Processing}, vol.~22, no.~4, pp. 1382--1394, April
  2013.

\bibitem[Ebrahimi and Vrscay(2007)]{ebrahimi_2007_solving}
M.~Ebrahimi and E.~R. Vrscay, ``Solving the inverse problem of image zooming
  using “self-examples”,'' in \emph{Image analysis and Recognition}.\hskip
  1em plus 0.5em minus 0.4em\relax Springer, 2007, pp. 117--130.

\bibitem[Egiazarian and Katkovnik(2015)]{egiazarian_2015_single}
K.~Egiazarian and V.~Katkovnik, ``Single image super-resolution via bm3d sparse
  coding,'' in \emph{EUSIPCO}, 2015.

\bibitem[Facchinei and Kanzow(2007)]{facchinei_2007_generalized}
F.~Facchinei and C.~Kanzow, ``Generalized nash equilibrium problems,''
  \emph{4OR}, vol.~5, no.~3, pp. 173--210, sep 2007.

\bibitem[Fernandez-Granda and Candes(2013)]{fernandez-granda_2013_super}
C.~Fernandez-Granda and E.~Candes, ``Super-resolution via transform-invariant
  group-sparse regularization,'' in \emph{IEEE International Conference on
  Computer Vision}, 2013, pp. 3336--3343.

\bibitem[Freedman and Fattal(2011)]{freedman_2011_image}
G.~Freedman and R.~Fattal, ``Image and video upscaling from local
  self-examples,'' \emph{ACM Transactions on Graphics (TOG)}, vol.~30, no.~2,
  p.~12, 2011.

\bibitem[Freeman et~al.(2002)Freeman, Jones, and Pasztor]{freeman_2002_example}
W.~T. Freeman, T.~R. Jones, and E.~C. Pasztor, ``Example-based
  super-resolution,'' \emph{Computer Graphics and Applications, IEEE}, vol.~22,
  no.~2, pp. 56--65, 2002.

\bibitem[Glasner et~al.(2009)Glasner, Bagon, and Irani]{glasner_2009_super}
D.~Glasner, S.~Bagon, and M.~Irani, ``Super-resolution from a single image,''
  in \emph{Intern. Conf. on Computer Vision}, 2009, pp. 349--356.

\bibitem[Guleryuz(2006)]{guleryuz_2006_nonlinear_a}
O.~G. Guleryuz, ``Nonlinear approximation based image recovery using adaptive
  sparse reconstructions and iterated denoising-part i: theory,'' \emph{IEEE
  Transactions on Image Processing}, vol.~15, no.~3, pp. 539--554, March 2006.

\bibitem[Haris et~al.(2016)Haris, Widyanto, and Nobuhara]{haris_2016_first}
M.~Haris, M.~R. Widyanto, and H.~Nobuhara, ``First-order derivative-based
  super-resolution,'' \emph{Signal, Image and Video Processing}, vol.~11,
  no.~1, pp. 1--8, mar 2016.

\bibitem[Huang et~al.(2015)Huang, Singh, and Ahuja]{huang_2015_single}
J.-B. Huang, A.~Singh, and N.~Ahuja, ``Single image super-resolution from
  transformed self-exemplars,'' in \emph{IEEE Conference on Computer Vision and
  Pattern Recognition)}, 2015.

\bibitem[Kim et~al.(2016{\natexlab{a}})Kim, Kwon~Lee, and
  Mu~Lee]{kim_2016_accurate}
J.~Kim, J.~Kwon~Lee, and K.~Mu~Lee, ``Accurate image super-resolution using
  very deep convolutional networks,'' in \emph{The IEEE Conference on Computer
  Vision and Pattern Recognition (CVPR Oral)}, Jun. 2016.

\bibitem[Kim et~al.(2016{\natexlab{b}})Kim, Kwon~Lee, and
  Mu~Lee]{kim_2016_deeply}
------, ``Deeply-recursive convolutional network for image super-resolution,''
  in \emph{The IEEE Conference on Computer Vision and Pattern Recognition (CVPR
  Oral)}, Jun. 2016.

\bibitem[Kim and Kim(2016)]{kim_2016_discrete}
J.~Kim and C.~Kim, ``Discrete feature transform for low-complexity single-image
  super-resolution,'' in \emph{2016 Asia-Pacific Signal and Information
  Processing Association Annual Summit and Conference ({APSIPA})}.\hskip 1em
  plus 0.5em minus 0.4em\relax Institute of Electrical and Electronics
  Engineers ({IEEE}), dec 2016.

\bibitem[Liu et~al.(2016)Liu, Wang, Wen, Yang, Han, and Huang]{liu_2016_robust}
D.~Liu, Z.~Wang, B.~Wen, J.~Yang, W.~Han, and T.~S. Huang, ``Robust single
  image super-resolution via deep networks with sparse prior,'' \emph{{IEEE}
  Transactions on Image Processing}, vol.~25, no.~7, pp. 3194--3207, jul 2016.

\bibitem[Salvador and P\'erez-Pellitero(2015)]{salvador_2015_naive}
J.~Salvador and E.~P\'erez-Pellitero, ``{Naive {Bayes} Super-Resolution
  Forest},'' in \emph{IEEE Int. Conf. on Computer Vision}, 2015.

\bibitem[Schulter et~al.(2015)Schulter, Leistner, and
  Bischof]{schulter_2015_fast}
S.~Schulter, C.~Leistner, and H.~Bischof, ``Fast and accurate image upscaling
  with super-resolution forests,'' in \emph{IEEE Conference on Computer Vision
  and Pattern Recognition}, 2015, pp. 3791--3799.

\bibitem[Shi and Qi(2016)]{shi_2016_lowrank}
J.~Shi and C.~Qi, ``Low-rank sparse representation for single image
  super-resolution via self-similarity learning,'' in \emph{2016 IEEE
  International Conference on Image Processing (ICIP)}, Sept 2016, pp.
  1424--1428.

\bibitem[Sidike et~al.(2017)Sidike, Krieger, Alom, Asari, and
  Taha]{sidike_2017_fast}
P.~Sidike, E.~Krieger, M.~Z. Alom, V.~K. Asari, and T.~Taha, ``A fast
  single-image super-resolution via directional edge-guided regularized extreme
  learning regression,'' \emph{Signal, Image and Video Processing}, jan 2017.

\bibitem[Singh and Ahuja(2014)]{singh_2014_sub}
A.~Singh and N.~Ahuja, ``Sub-band energy constraints for self-similarity based
  super-resolution,'' in \emph{International Conference on Pattern
  Recognition}, 2014, pp. 4447--4452.

\bibitem[Singh et~al.(2014)Singh, Porikli, and Ahuja]{singh_2014_super}
A.~Singh, F.~Porikli, and N.~Ahuja, ``Super-resolving noisy images,'' in
  \emph{IEEE Conference on Computer Vision and Pattern Recognition}.\hskip 1em
  plus 0.5em minus 0.4em\relax IEEE, 2014, pp. 2846--2853.

\bibitem[Suetake et~al.(2008)Suetake, Sakano, and Uchino]{suetake_2008_image}
N.~Suetake, M.~Sakano, and E.~Uchino, ``Image super-resolution based on local
  self-similarity,'' \emph{Optical review}, vol.~15, no.~1, pp. 26--30, 2008.

\bibitem[Sun et~al.(2011)Sun, Xu, and Shum]{sun_2011_gradient}
J.~Sun, Z.~Xu, and H.-Y. Shum, ``Gradient profile prior and its applications in
  image super-resolution and enhancement,'' \emph{IEEE Transactions on Image
  Processing}, vol.~20, no.~6, pp. 1529--1542, 2011.

\bibitem[Tang and Shao(2017)]{tang_2017_pairwise}
Y.~Tang and L.~Shao, ``Pairwise operator learning for patch-based single-image
  super-resolution,'' \emph{{IEEE} Transactions on Image Processing}, vol.~26,
  no.~2, pp. 994--1003, feb 2017.

\bibitem[Timofte et~al.(2013)Timofte, De, and Van~Gool]{timofte_2013_anchored}
R.~Timofte, V.~De, and L.~Van~Gool, ``Anchored neighborhood regression for fast
  example-based super-resolution,'' in \emph{IEEE International Conf. on
  Computer Vision}, 2013, pp. 1920--1927.

\bibitem[Timofte et~al.(2014)Timofte, De~Smet, and
  Van~Gool]{timofte_2014_adjusted}
R.~Timofte, V.~De~Smet, and L.~Van~Gool, ``A+: Adjusted anchored neighborhood
  regression for fast super-resolution,'' in \emph{Computer Vision--ACCV
  2014}.\hskip 1em plus 0.5em minus 0.4em\relax Springer, 2014, pp. 111--126.

\bibitem[Timofte et~al.(2016)Timofte, Rothe, and
  Van~Gool]{timofte_2016_improved}
R.~Timofte, R.~Rothe, and L.~Van~Gool, ``Seven ways to improve example-based
  single image super resolution,'' in \emph{The IEEE Conf. on Computer Vision
  and Pattern Recognition}, June 2016.

\bibitem[Wang et~al.(2016)Wang, Shao, Ge, Li, and Huang]{wang_2016_return}
F.~Wang, W.-Z. Shao, Q.~Ge, H.-B. Li, and L.-L. Huang, ``Return of
  reconstruction-based single image super-resolution: A simple and accurate
  approach,'' in \emph{2016 9th International Congress on Image and Signal
  Processing, {BioMedical} Engineering and Informatics ({CISP}-{BMEI})}.\hskip
  1em plus 0.5em minus 0.4em\relax Institute of Electrical and Electronics
  Engineers ({IEEE}), oct 2016.

\bibitem[Wei and Dragotti(2016)]{wei_2016_freshfri}
X.~Wei and P.~L. Dragotti, ``{FRESH}{\textemdash}{FRI}-based single-image
  super-resolution algorithm,'' \emph{{IEEE} Transactions on Image Processing},
  vol.~25, no.~8, pp. 3723--3735, aug 2016.

\bibitem[Yang et~al.(2010)Yang, Wright, Huang, and Ma]{yang_2010_image}
J.~Yang, J.~Wright, T.~S. Huang, and Y.~Ma, ``Image super-resolution via sparse
  representation,'' \emph{IEEE Transactions on Image Processing}, vol.~19,
  no.~11, pp. 2861--2873, 2010.

\bibitem[Yang et~al.(2012)Yang, Wang, Lin, Cohen, and Huang]{yang_2012_coupled}
J.~Yang, Z.~Wang, Z.~Lin, S.~Cohen, and T.~Huang, ``Coupled dictionary training
  for image super-resolution,'' \emph{IEEE Transactions on Image Processing},
  vol.~21, no.~8, pp. 3467--3478, 2012.

\bibitem[Yang and Wang(2013)]{yang_2013_self}
M.~C. Yang and Y.~C.~F. Wang, ``A self-learning approach to single image
  super-resolution,'' \emph{IEEE Transactions on Multimedia}, vol.~15, no.~3,
  pp. 498--508, April 2013.

\bibitem[Yao et~al.(2017)Yao, Zhang, Bao, Liu, and Zhang]{yao_2017_blending}
X.~Yao, Y.~Zhang, F.~Bao, Y.~Liu, and C.~Zhang, ``The blending interpolation
  algorithm based on image features,'' \emph{Multimedia Tools and
  Applications}, jan 2017.

\bibitem[Zeng et~al.(2017)Zeng, Yu, Wang, Li, and Tao]{zeng_2017_coupled}
K.~Zeng, J.~Yu, R.~Wang, C.~Li, and D.~Tao, ``Coupled deep autoencoder for
  single image super-resolution,'' \emph{{IEEE} Transactions on Cybernetics},
  vol.~47, no.~1, pp. 27--37, jan 2017.

\bibitem[Zeyde et~al.(2012)Zeyde, Elad, and Protter]{zeyde_2012_single}
R.~Zeyde, M.~Elad, and M.~Protter, ``On single image scale-up using
  sparse-representations,'' in \emph{Curves and Surfaces}.\hskip 1em plus 0.5em
  minus 0.4em\relax Springer, 2012, pp. 711--730.

\bibitem[Zhang et~al.(2016)Zhang, Wang, Zuo, Zhang, and
  Zhang]{zhang_2016_joint}
K.~Zhang, B.~Wang, W.~Zuo, H.~Zhang, and L.~Zhang, ``Joint learning of multiple
  regressors for single image super-resolution,'' \emph{{IEEE} Signal
  Processing Letters}, vol.~23, no.~1, pp. 102--106, jan 2016.

\bibitem[Zhu et~al.(2014)Zhu, Zhang, and Yuille]{zhu_2014_single}
Y.~Zhu, Y.~Zhang, and A.~L. Yuille, ``Single image super-resolution using
  deformable patches,'' in \emph{IEEE Conference on Computer Vision and Pattern
  Recognition}, 2014, pp. 2917--2924.

\end{thebibliography}
